\documentclass{article}

\usepackage[final,nonatbib]{neurips_2020}

\usepackage[english]{babel}

\usepackage{makecell}
\usepackage{colortbl}
\usepackage{xcolor}

\definecolor{applegreen}{rgb}{0.55, 0.71, 0.0}
\definecolor{darkcyan}{rgb}{0.0, 0.55, 0.55}
\definecolor{darkpastelgreen}{rgb}{0.01, 0.75, 0.24}

\usepackage{pdfpages}
\usepackage[utf8]{inputenc} 
\usepackage[T1]{fontenc}    
\usepackage[breaklinks=false,letterpaper=true,colorlinks,citecolor=darkpastelgreen,bookmarks=false]{hyperref}       
\usepackage{url}            
\usepackage{booktabs}       
\usepackage{amsfonts,amsmath}       
\usepackage{nicefrac}       
\usepackage{microtype}      
\usepackage{graphicx}

\usepackage{algorithm}
\usepackage[noend]{algpseudocode}

\usepackage{array,multirow}

\usepackage{booktabs} 

\usepackage{float}
\floatstyle{plaintop}
\restylefloat{table}

\usepackage[
backend=biber,
maxcitenames=50,
maxnames=50,
style=numeric,
citestyle=numeric
]{biblatex}

\usepackage{adjustbox}
\usepackage[misc]{ifsym}

\usepackage{rotating}

\usepackage{xspace}

\makeatletter
\DeclareRobustCommand\onedot{\futurelet\@let@token\@onedot}
\def\@onedot{\ifx\@let@token.\else.\null\fi\xspace}
\def\eg{\emph{e.g}\onedot} 
\def\ie{\emph{i.e}\onedot}

\def\aka{\emph{a.k.a}\onedot}

\def\etal{\emph{et al}\onedot}
\makeatother

\newcommand{\navref}[2]{{\ref{#1}{\color{red}{\hyperref[#1]{#2}}}}}

\newcommand{\expectation}{\mathop{\mathbb{E}}}

\newcommand{\navcomment}[1]{{\color{blue}$\triangleright$ #1}}

\DeclareMathOperator*{\argmax}{arg\,max}

\title{Your Classifier can Secretly Suffice \\ Multi-Source Domain Adaptation}

\author{%
  Naveen Venkat ~~~~~ Jogendra Nath Kundu ~~~~~ Durgesh Kumar Singh \\ \textbf{Ambareesh Revanur} ~~~~~ \textbf{R. Venkatesh Babu} \\
  Video Analytics Lab, Indian Institute of Science, Bangalore \\
  {\small Corresponding author: \texttt{nav.naveenvenkat@gmail.com}} \\
}

\begin{document}

\maketitle

\begin{abstract}
Multi-Source Domain Adaptation (MSDA) deals with the transfer of task knowledge from multiple labeled source domains to an unlabeled target domain, under a domain-shift. Existing methods aim to minimize this domain-shift using auxiliary distribution alignment objectives. In this work, we present a different perspective to MSDA wherein deep models are observed to implicitly align the domains under label supervision. Thus, we aim to utilize implicit alignment without additional training objectives to perform adaptation. To this end, we use pseudo-labeled target samples and enforce a classifier agreement on the pseudo-labels, a process called Self-supervised Implicit Alignment (SImpAl). We find that SImpAl readily works even under category-shift among the source domains. Further, we propose classifier agreement as a cue to determine the training convergence, resulting in a simple training algorithm. We provide a thorough evaluation of our approach on five benchmarks, along with detailed insights into each component of our approach\footnote{Project page: \textbf{\url{https://sites.google.com/view/simpal}}}.


\end{abstract}

\section{Introduction}

The task of supervised learning for classification is based on the assumption that
the training data and the testing data are sampled from the same distributions. Thus, supervised learning methods achieve state-of-the-art results when evaluated on popular benchmarks such as ImageNet~\cite{imagenet}. However, when such models are deployed in real-world, they yield sub-optimal results due to the inherent distribution-shift (domain-shift~\cite{torralba2011unbiased}) between the training data and the real-world environment (\aka the target domain). While it is possible to obtain unlabeled samples from the target domain in most cases, the huge costs of data annotation prohibit the creation of a reliable labeled training dataset. To this end, Unsupervised Domain Adaptation (DA) methods have been proposed that aim to transfer knowledge from a labeled "source" dataset to an unlabeled "target" dataset under a domain-shift.


A popular strategy in Unsupervised DA is to learn the task-specific knowledge using supervision from the labeled source dataset, while learning a domain-invariant latent space where the features across the source and the target domains align. Such an alignment is enforced using statistical discrepancy minimization schemes \cite{JDA_Dist_Match,gong2012geodesic,pan2010domain_tca,peng2019moment,sun2016deepcoral} or via an adversarial objective \cite{ganin2016domain,long2018conditional,tzeng2017adversarial,dctn,zhao2018adversarial}, or by employing domain-specific transformations~\cite{dsbn,adabn,featurewhiteningandconsensusloss}. This alignment minimizes the domain-shift in the latent space, and improves the target generalization.
However, the performance of Single-Source Domain Adaptation (SSDA) methods is usually determined by the choice of the source dataset~\cite{kundu2020towards}.

\begin{figure}[t]
    \centering
    \includegraphics[width=1\linewidth]{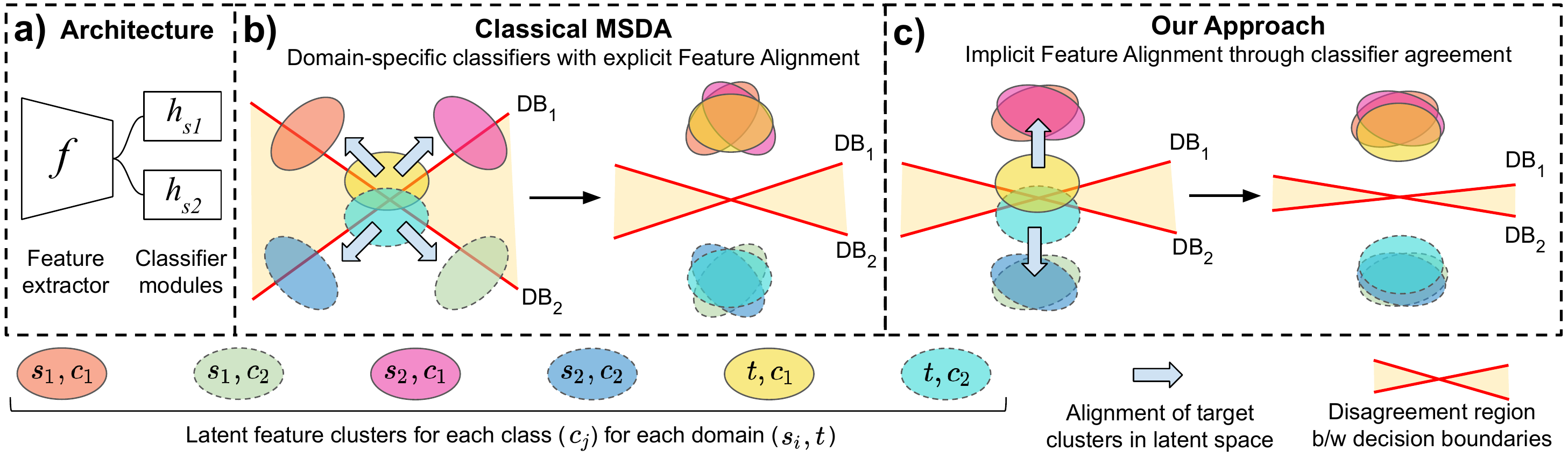}
    \vspace{-5mm}
    \caption{\small An illustration of the proposed concept for two-class ($c_1, c_2$) classification, with two labeled source domains ($s_1, s_2$) and one unlabeled target domain ($t$). 
    Best viewed in color. 
    \textbf{(a) Architecture.} Several works employ a shared feature extractor ($f$), and source-specific classifier modules.
    \textbf{(b) Classical MSDA.} Prior works employ source-specific classifiers that learn distinct (domain-specific) decision boundaries (denoted as $\text{DB}_i$). This results in a large discrepancy in the classifier predictions (region shaded in yellow). Thus, an auxiliary feature alignment loss is required for improving the classifier predictions. \textbf{(c) Our approach.} We enforce classifier agreement on source-domain samples leading to an \textit{implicit alignment} of latent features. Further, imposing an agreement on the pseudo-labeled target samples improves the generalization to the target domain. }
    \label{fig:basic_overview}
\end{figure}

Recently, Multi-Source Domain Adaptation (MSDA) \cite{mansour2012multiple,zhao2020multi} has garnered interest wherein multiple labeled source domains are used to transfer the task knowledge to the unlabeled target domain.
A common approach \cite{mixtureofexperts,peng2019moment,dctn} is to learn a shared feature extractor, along with domain-specific classifier modules (Fig.~\navref{fig:basic_overview}{a}), which yield an ensemble prediction for the target samples. 
However, an additional challenge in MSDA is to tackle the domain-shift and category-shift~\cite{dctn} between each pair of source-domains (Fig.~\navref{fig:basic_overview}{b}). To this end, auxiliary losses are enforced encouraging the model to learn domain invariant but class-discriminative representations. Ultimately, an appropriate alignment of all the domains in the latent space \cite{peng2019moment} improves the generalization on the target domain (Fig.~\navref{fig:basic_overview}{b}). 



\begin{figure}[!t]
    \centering
    \includegraphics[width=1\linewidth]{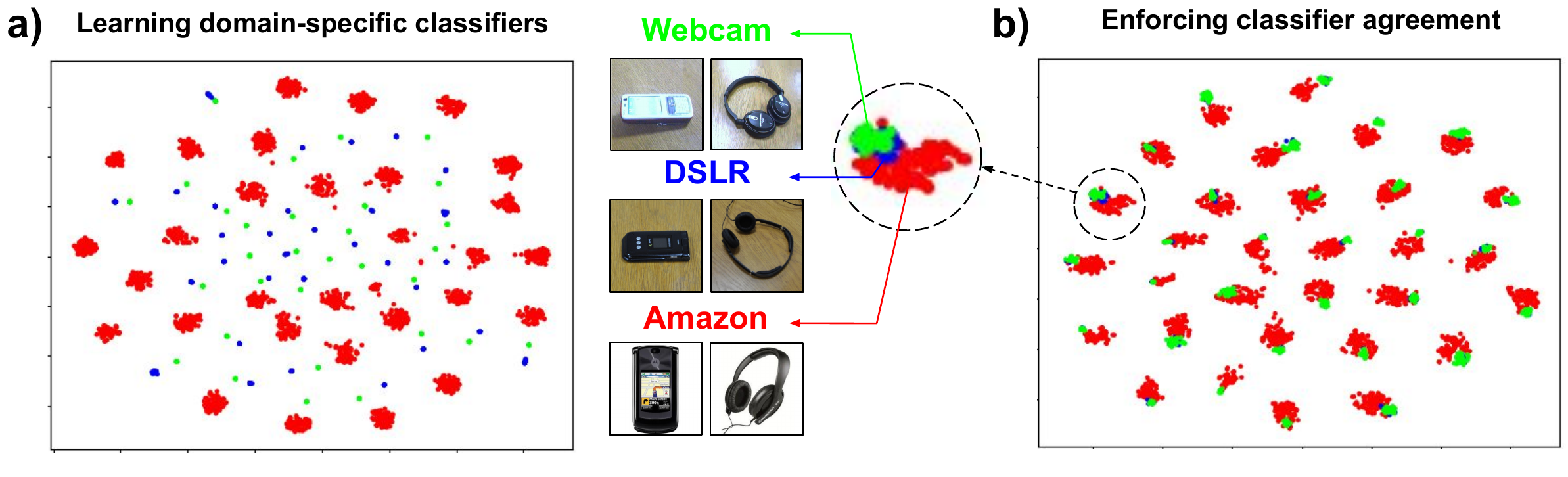}
    \vspace{-6mm}
    \caption{\small t-SNE plot of the features at the pre-classifier space, showing the feature distribution after learning a fully supervised model. Best viewed in color. \textbf{(a) Learning domain-specific classifiers.} We train a model (with ResNet-50 \cite{he2016deep_resnet} backbone) with full label supervision from all the three domains on Office-31~\cite{office}, while keeping the classifier heads unique to each domain. Although we find that class discrimination is achieved, each domain forms separate sub-clusters, and does not align in the latent space.
    \textbf{(b) Enforcing classifier agreement.} Instead of learning domain-specific classifiers, we enforce the classifiers to agree upon the labels for all the samples. We observe that all the domains tend to align, without enforcing an explicit alignment objective, even under a domain-shift. We aim to leverage this inductive bias of deep models, to perform adaptation. 
    }
    \label{fig:motivation_tsne}
    \vspace{-1mm}
\end{figure}

In this work, we approach the MSDA problem from a different perspective. 
Since deep models are known to capture rich transferable representations~\cite{long2015learning,oquab2014learning,howtransferable}, we ask, \textit{is an auxiliary feature alignment loss really necessary}?
The motivation stems from the observation that deep models exhibit a strong \textit{inductive bias} to implicitly align the latent features under supervision. This is demonstrated in Fig.~\ref{fig:motivation_tsne}. Following the prior approaches~\cite{peng2019moment,dctn}, we train domain-specific classifiers (Fig.~\navref{fig:basic_overview}{b}) and observe that the domains do not align in the latent space (Fig.~\navref{fig:motivation_tsne}{a}), which calls for an explicit feature alignment loss. However, when we enforce a classifier agreement on the class label for each input instance (Fig.~\navref{fig:motivation_tsne}{b}), we find that the domains tend to align, without requiring an explicit alignment loss. 


This motivates us to further explore \textit{implicit alignment} of latent features for MSDA. We aim to leverage the labeled data from multiple source domains, and the multi-classifier setup (Fig.~\navref{fig:basic_overview}{a}) employed in MSDA to perform alignment, without incorporating auxiliary components such as a domain discriminator \cite{dctn,zhao2018adversarial}. In contrast to learning domain-specific classifier modules, we enforce an agreement among the classifiers (Fig.~\navref{fig:basic_overview}{c}) to align the domains in the latent space.

Since the target domain is unlabeled, we resort to the class labels predicted by the model being trained (\aka pseudo-labels~\cite{lee2013pseudo}). The adaptation step encourages the classifiers to agree upon these pseudo-labels which enables alignment of the target features with the source features that have classifier agreement owing to label supervision.
Accordingly, we name the approach as \textbf{S}elf-supervised \textbf{Imp}licit \textbf{Al}ignment, abbreviated as SImpAl (pronounced "simple"). We observe that even under category-shift, implicit alignment can be leveraged to align the shared categories, without requiring additional components (\eg fine-grained alignment~\cite{cao2018partial1,kundu2020class,mada}, adversarial discriminator~\cite{dctn}) or cumbersome training strategies (\eg to handle arbitrary category-shifts~\cite{kundu_cvpr_2020,dctn,UDA_2019_CVPR}). 
We also find that classifier agreement can be leveraged as a cue to determine adaptation convergence.

To summarize, we demonstrate successful MSDA by leveraging implicit alignment exhibited by deep classifiers, corroborating the potential for designing simple and effective adaptation algorithms. We conduct extensive evaluation of our approach over five benchmark datasets, with two popular CNN backbone models (ResNet-50, ResNet-101 \cite{he2016deep_resnet}) and derive insights from the empirical analysis.



\section{Related Work}
\label{sec:related_work}

Here, we briefly review the related works and refer the reader to \cite{zhao2020multi} for an extensive survey. 

\textbf{a) Single-Source Domain Adaptation (SSDA).} Motivated by the seminal work by Ben-David \etal \cite{ben2010theory,ben2007analysis}, a large number of SSDA methods \cite{dsbn,NormalAdaptfirstBackprop,ganin2016domain,gong2012geodesic,sta_open_set,long2015learning,long2016unsupervised} have been proposed, that aim to learn domain-agnostic but class-discriminative representations. Inspired by the GAN framework~\cite{goodfellow2014generative}, a popular strategy is to employ adversarial learning~\cite{hoffman2018cycada,adadepth,saito2018open,sankaranarayanan2018generate,tzeng2015simultaneous,tzeng2017adversarial,tzeng2014deep} that aims to confuse a domain-discriminator, thereby aligning the latent features of the domains. Saito \etal \cite{saito2018maximum} formulate an adversarial objective employing classifier discrepancy. In contrast, we aim to study a simpler approach which circumvents the training difficulties encountered in adversarial learning paradigms. Recently, consistency based regularizers \cite{chen2019crdoco,kundu2019_um_adapt,murez2018image,adadepth} were proposed for domain adaptation. In our work, classifier agreement can be interpreted as a form of consistency at the output space which acts both as an implicit regularizer and as a means to perform latent space alignment for adaptation.


\textbf{b) Multi-Source Domain Adaptation (MSDA).} Several methods \cite{mixtureofexperts,peng2019moment,dctn,zhu2019aligning_mfsan} learn domain-specific classifier modules and obtain a weighted ensemble prediction for the target samples, motivated by the distribution weighted combining rule \cite{hoffman2018algorithms,mansour2009domain,mansour2012multiple}. Zhe \etal \cite{zhu2019aligning_mfsan} employ an alignment loss between each source-target pair in domain-specific feature spaces. In addition, Peng \etal \cite{peng2019moment} align each pair of source domains using kernel based moment matching and also propose a variant based on adversarial learning \cite{saito2018maximum}. Xu \etal employ multiple domain discriminators to achieve latent space alignment. In this work, we aim to explore a simple adaptation scheme that leverages implicit alignment in deep models. As a result, our approach is applicable even under category-shift among the source domains, while most prior methods \cite{mixtureofexperts,peng2019moment,zhu2019aligning_mfsan} consider only a shared category set.

\textbf{c) Self-training methods.} Pseudo-labeling~\cite{lee2013pseudo} is a popular semi-supervised learning approach where "pseudo" class labels are assigned to unlabeled samples, typically using classifier confidence~\cite{chen2011co,saito_asymmetric_tritraining,dctn,zou2019confidence,zou2018unsupervised} or nearest neighbor assignment~\cite{kundu2020class,pan2019transferrable,saito2020universal,zhang2019category}, while the model is retrained using such samples. Confidence thresholding~\cite{li2019bidirectional,saito_asymmetric_tritraining,dctn} is commonly applied to minimize the noise in pseudo-labels. This introduces a sensitive threshold hyperparameter, requiring labeled target samples or domain expertise for precise tuning. Works such as Zou \etal \cite{zou2019confidence,zou2018unsupervised}, Li \etal \cite{li2019bidirectional} and Chen \etal \cite{chen2019crdoco} propose various regularizers to improve pseudo-label predictions. Xu \etal \cite{dctn} incorporate an adversarial alignment loss to mitigate the performance degradation arising from noisy pseudo-labels. In contrast, we aim to exploit classifier agreement to perform adaptation and improve the reliability of pseudo-labels without incorporating additional hyperparameters.
\section{Self-supervised Implicit Alignment (SImpAl)}
\label{sec:method}



\textbf{Notations.} 
Let $\mathcal{X}$ and $\mathcal{Y}$ denote the input and the output spaces. We consider $n_d$ labeled source domain datasets $\{{\mathcal{D}_{s_i}}\}_{i=1}^{n_d}$, where ${\mathcal{D}_{s_i}} = \{(\mathbf{x}_{s_i}^{c_j}, y_{s_i}^{c_j}) \in \mathcal{X} \times \mathcal{Y} \}$ and a single unlabeled target domain dataset ${\mathcal{D}_t} = \{\mathbf{x}_t \in \mathcal{X}\}$. 
Each source domain has a label-set $\mathcal{C}_{s_i}$, and the target label-set is defined as $\mathcal{C} = \cup_{i=1}^{n_d} \mathcal{C}_{s_i}$ with $n_c$ classes. 
We learn a deep neural network model having a CNN based feature extractor $f : \mathbb{R}^{224 \times 224 \times 3} \rightarrow \mathbb{R}^{256}$, and $n_d$ classifier modules $h : \mathbb{R}^{256} \rightarrow \mathbb{R}^{n_d \times n_c}$. For convenience, we denote the output of the network as a matrix $\mathbf{M} = h \circ f (\mathbf{x})$, where $\circ$ represents function composition. $\mathbf{M}$ is obtained by stacking the logits produced by each classifier (see Fig.~\ref{fig:architecture}). 

\textbf{Overview.} As is conventional in the MSDA methods~\cite{mixtureofexperts,dctn}, the multi-classifier setup is treated as an ensemble of diverse classifiers, and the class probabilities are obtained through a convex combination of each classifier's prediction. The model is first trained with the categorical cross-entropy loss imposed on the combined data from all source domains. After a "warm-start", we introduce pseudo-labeled target samples into the training process. The adaptation is performed by enforcing the classifiers to agree on these pseudo-labels. We now describe the approach in detail.


\subsection{Warm-start with source domains}
\label{subsec:warm_start}

To adapt the network to the target domain, we use pseudo-labeled target samples. Thus, we first aim to achieve a reliability in pseudo-labels by training the model on all source domains. We call this as the warm-start process, which is performed as follows.

\begin{figure}[!t]
    \centering
    \includegraphics[width=1\linewidth]{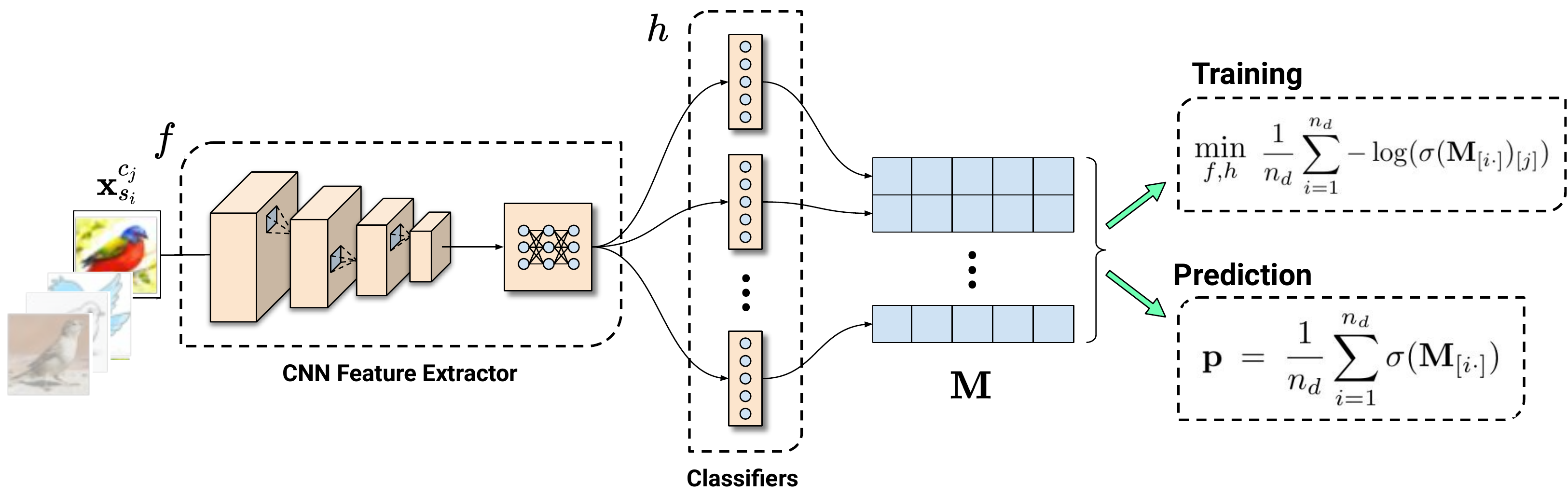}
    \vspace{-5mm}
    \caption{\small Architecture of the proposed approach. The  network contains a common feature extractor $f$ having a CNN backbone followed by fully-connected layers. The multi-classifier module $h$ contains $n_d$ classifiers.}
    \label{fig:architecture}
    \vspace{-2mm}
\end{figure}

\textbf{a) Learning with source domains.} For each source-domain instance $\mathbf{x}_{s_i}^{c_j}$, we obtain the output matrix $\mathbf{M} = h \circ f(\mathbf{x}_{s_i}^{c_j})$ (see Fig.~\ref{fig:architecture}) and define the class probability vector $\mathbf{p}$ as a convex combination of the probabilities assigned by each classifier,

\vspace{-3mm}
\begin{equation}
    \label{eq:ensemble_prediction}
    \mathbf{p} ~=~ \frac{1}{n_d} \sum_{i=1}^{n_d} \sigma(\mathbf{M}_{[i \cdot]})
\end{equation}
\vspace{-1mm}

where, $\mathbf{M}_{[i \cdot]}$ represents the $i^{\text{th}}$ row vector of the matrix $\mathbf{M}$ (\ie the logits of the $i^{\text{th}}$ classifier), and $\sigma(\mathbf{v})_{[j]} = \exp{(\mathbf{v}_{[j]})} / \sum_{j'=1}^{n_c} \exp{(\mathbf{v}_{[j']})}$ is the $\operatorname{softmax}$ function. Treating $\mathbf{p}$ as the class probability vector, 
we minimize the categorical cross entropy loss ($l_{ce}$) using the labeled source samples,

\vspace{-4mm}
\begin{equation}
    \label{eq:ensemble_ce}
    l_{ce}(\mathbf{p}, y_{s_i}^{c_j}) ~~=~~ -\operatorname{log}(\mathbf{p}_{[j]}) ~~=~~ -\operatorname{log}\Big(\frac{1}{n_d} \sum_{i=1}^{n_d} \sigma(\mathbf{M}_{[i \cdot]})_{[j]}\Big) ~~\leq~~ \frac{1}{n_d} \sum_{i=1}^{n_d} - \operatorname{log}(\sigma(\mathbf{M}_{[i \cdot]})_{[j]})
\end{equation}
\vspace{-2mm}



The last term in Eq.~\ref{eq:ensemble_ce} represents an upper bound for the categorical cross-entropy loss of the ensemble, and is obtained by applying the Jensen's inequality for convex functions. We consider the formulation in Eq.~\ref{eq:ensemble_ce} to drive the classifiers to agree upon the label $y_{s_i}^{c_j}$ for $\mathbf{x}_{s_i}^{c_j}$. Thus, the training objective is,

\vspace{-3mm}
\begin{equation}
    \label{eq:our_mirror_loss}
    \min_{f, h} ~~~ \expectation_{\mathcal{D} \in \{\mathcal{D}_{s_{i'}}\}_{i'=1}^{n_d}} ~ \expectation_{(\mathbf{x}_{s_i}^{c_j}, y_{s_i}^{c_j}) \in \mathcal{D}} ~ \frac{1}{n_d} \sum_{i=1}^{n_d} - \operatorname{log}(\sigma(\mathbf{M}_{[i \cdot]})_{[j]})
\end{equation}
\vspace{-1mm}

The objective in Eq.~\ref{eq:our_mirror_loss} is minimized by mini-batch stochastic optimization. Each mini-batch contains an equal number of samples from each source domain. In practice, each classifier is given a distinct random initialization, and is trained with the same set of training samples at each mini-batch. Intuitively, this process gradually enables a higher degree of similarity among the classifiers (Fig.~\navref{fig:basic_overview}{c}) through an agreement in the predicted class labels for source samples. Note that, both the feature extractor $f$ and the multi-classifier module $h$ are shared across all source domains. This step provides a warm-start to introduce pseudo-labeled target samples into the training.

\textbf{b) Determining the convergence of warm-start.} 
The next question we address is, \textit{how to tell if a model is trained sufficiently for the target domain?} 
Intuitively, we would like to train the model until there is a saturation in the target (pseudo-label) accuracy. However, with unlabeled target samples measuring the pseudo-label accuracy is out of bounds. Thus, we propose the classifier agreement as a criterion to determine the convergence.
The classifier agreement for an instance $\mathbf{x}$ is defined as,

\vspace{-3mm}
\begin{gather}
    a(\mathbf{x}, f, h) = {\prod}_{i \neq i'} I(~\argmax_{j\in\mathcal{C}} \mathbf{M}_{[ij]} ~=~ \argmax_{j\in\mathcal{C}} \mathbf{M}_{[i'j]}~)
\end{gather}
\vspace{-2mm}

where $\mathbf{M} = h \circ f (\mathbf{x})$, and $I(\cdot)$ is the indicator function that returns $1$ when the condition is true, else returns $0$.
Intuitively, when each classifier predicts the same class for a given sample $\mathbf{x}$, we say that the classifiers "agree". Thus, $a(\mathbf{x}, f, h) = 1$ when classifiers agree, and $a(\mathbf{x}, f, h) = 0$ otherwise. 

As we shall show in Sec.~\ref{subsec:analysis}, the target pseudo-label accuracy is higher whenever the classifiers agree~\cite{nguyen2019self,yu2019does}. Thus, classifier agreement is used to filter out target samples having a higher degree of noise in pseudo-labels. Further, we estimate the fraction of target samples for which there is an agreement in the class predictions among the classifiers. Thus, we define the target agreement rate as,

\vspace{-3mm}
\begin{equation}
    \label{eq:agreement_rate}
    A({\mathcal{D}_t}, f, h) = \frac{1}{|{\mathcal{D}_t}|} \sum_{\mathbf{x}_t \in {\mathcal{D}_t}} a(\mathbf{x}_t, f, h)
\end{equation}
\vspace{-3mm}

We hypothesize that the performance on target samples attains a saturation when the agreement rate converges. Thus, we determine the warm-start interval based on the convergence of $A(\cdot)$.

\subsection{Introducing target data}
\label{subsec:target_tuning}

After the warm-start, we introduce target samples into the training process. The pseudo-labels are obtained from the classifier predictions as in Eq.~\ref{eq:ensemble_prediction}, \ie 
$y_t^{c_j} = \argmax_{j'} \frac{1}{n_d} \sum_{i=1}^{n_d} \sigma(\mathbf{M}_{[ij']})$.

 
We consider the following strategy for pseudo-labeling.
To begin with, we select only those target samples for which there is a classifier agreement, since the labels are seen to be more accurate for such samples (verified in Sec.~\ref{subsec:analysis}). Thus, we obtain a subset ${\mathcal{D}_t}' = \{(\mathbf{x}_t, y_t^{c_j}) ~|~ \mathbf{x}_t \in {\mathcal{D}_t}, ~a(\mathbf{x}_t, f, h) = 1\}$. Secondly, inspired by curriculum learning~\cite{bengio2009curriculum,zou2018unsupervised} we form an easy-to-hard sampling strategy for ${\mathcal{D}_t}'$.
For this purpose, we obtain the average classifier margin as a weight for each target instance,

\vspace{-3mm}
\begin{equation}
    \label{eq:margin}
    w(\mathbf{x}_t, f, h) ~=~ \frac{1}{n_d} \sum_{i=1}^{n_d} (\mathbf{M}_{[i j]} - \mathbf{M}_{[i j']})
\end{equation}
\vspace{-2mm}

where $j$ an $j'$ correspond to the indices of the highest and the second highest logit. Intuitively, $w$ measures a form of confidence in prediction. Target samples that are farther from the decision boundaries receive a higher $w$ (see Fig.~\navref{fig:target_migration}{c} for the geometrical interpretation). We show in Sec. \ref{subsec:analysis} that, in general, samples with a higher $w$ are more likely to possess correct pseudo-labels. Thus, target samples are sorted based on $w$, and are fed to the training pipeline in the decreasing order of $w$. Finally, the pseudo-labels are updated every $n_e$ epochs on ${\mathcal{D}_t}'$. With this strategy, we formalize the training objective for adaptation using the target samples as,

\vspace{-3mm}
\begin{equation}
    \label{eq:our_target_mirror_loss}
    \min_{f, h} ~~~ \expectation_{(\mathbf{x}_t, y_t^{c_j}) \in {\mathcal{D}_t}' } ~ \frac{1}{n_d} \sum_{i=1}^{n_d} - \operatorname{log}(\sigma(\mathbf{M}_{[i \cdot]})_{[j]})
\end{equation}
\vspace{-2mm}


After introducing the target samples from ${\mathcal{D}_t}'$, we train on both source and target samples, in alternate mini-batches, \ie we minimize the objectives in Eq.~\ref{eq:our_mirror_loss} and Eq.~\ref{eq:our_target_mirror_loss} in alternate mini-batches. Finally, the network is trained until the target agreement rate $A$ shows convergence. This enables a simple and effective adaptation pipeline using implicit alignment. The algorithm is given in Algo.~{\color{red}1}.

\begin{algorithm}[!ht]
\label{algo}
\caption{SImpAl - Self-supervised Implicit Alignment}
\begin{algorithmic}[1]

\State \textbf{require:} Source datasets $\{{\mathcal{D}_{s_i}}\}_{i=1}^{n_d}$, Target dataset ${\mathcal{D}_t}$, Model $\{f, h\}$

\Statex



\While {$A({\mathcal{D}_t}, f, h)$ has not converged} \hfill \navcomment{Warm-start with source domains}

    \State Load a mini-batch of samples $(\mathbf{x}_{s_i}^{c_j}, y_{s_i}^{c_j})$ from each source dataset ${\mathcal{D}_{s_i}}$

    \State Update $\{f, h\}$ using the objective in Eq.~\ref{eq:our_mirror_loss}

\EndWhile

\Statex

\State Obtain pseudo-labeled target subset ${\mathcal{D}_t}' = \{(\mathbf{x}_t, y_t^{c_j}) ~|~ \mathbf{x}_t \in \mathcal{D}_t, ~a(\mathbf{x}_t, f, h) = 1\}$

\State Prepare ${\mathcal{D}_t}'$ by sorting the samples in descending order of $w(\mathbf{x}_t, f, h)$ (as in Eq.~\ref{eq:margin})

\Statex 

\While {$A({\mathcal{D}_t}, f, h)$ has not converged} \hfill \navcomment{Introducing target samples}
    
    \Statex

    \State Load a mini-batch of samples $(\mathbf{x}_{s_i}^{c_j}, y_{s_i}^{c_j})$ from each source dataset ${\mathcal{D}_{s_i}}$

    \State Update $\{f, h\}$ using the objective in Eq.~\ref{eq:our_mirror_loss}

    \Statex

    \State Load a mini-batch of samples $(\mathbf{x}_{t}, y_{t}^{c_j})$ from pseudo-labeled target subset ${\mathcal{D}_t}'$

    \State Update $\{f, h\}$ using the objective in Eq.~\ref{eq:our_target_mirror_loss}
    
    \Statex
    
    \If{ $n_e$ epochs on ${\mathcal{D}_t}'$ are completed } \hfill \navcomment{Periodically update pseudo-labels}
    
        \State Perform steps {\color{red} 5-6} to recompute ${\mathcal{D}_t}'$
    
    \EndIf

\EndWhile

\end{algorithmic}

\end{algorithm}

\section{Experiments}
\label{sec:experiments}

We present the results of our approach on five standard benchmark datasets - \textit{Office-Caltech}, \textit{ImageCLEF}, \textit{Office-31}, \textit{Office-Home} and the most challenging large-scale benchmark, \textit{DomainNet}.

\textbf{a) Prior Arts.} We compare against Deep Domain Confusion (DDC)~\cite{tzeng2014deep}, Deep Adaptation Network (DAN)~\cite{long2015learning}, Deep CORAL (D-CORAL)~\cite{sun2016deepcoral}, Reverse Gradient (RevGrad)~\cite{NormalAdaptfirstBackprop}, Residual Transfer Network (RTN)~\cite{long2016unsupervised}, Joint Adaptation Network (JAN)~\cite{long2017deepJAN}, Maximum Classifier Discrepancy (MCD)~\cite{saito2018maximum}, Manifold Embedded Distribution Alignment (MEDA)~\cite{wang2018visual}, Adversarial Discriminative Domain Adaptation (ADDA)~\cite{tzeng2017adversarial}, Deep Cocktail Network (DCTN)~\cite{dctn}, Moment Matching (M$^3$SDA)~\cite{peng2019moment} and Multiple Feature Space Adaptation Network (MFSAN)~\cite{zhu2019aligning_mfsan}. Specifically, DDC, RevGrad, ADDA, MCD, DCTN use an adversarial alignment objective to perform adaptation, RTN learns a residual function to bridge the distribution discrepancy, and DAN, MFSAN, D-CORAL, JAN, MEDA and M$^3$SDA employ a kernel based moment matching scheme to align the domains.

\textbf{b) Evaluation.} For \textit{ImageCLEF} and \textit{Office}-based datasets, we follow the evaluation protocol in MFSAN~\cite{zhu2019aligning_mfsan}, while for DomainNet, we follow the protocol used in M$^3$SDA~\cite{peng2019moment}. Three types of baselines are considered - 1) Single Best (SB) refers to the best single-source transfer results for the target domain, 2) Source Combine (SC) refers to the scenario where all sources are combined into a single source domain to perform SSDA, 3) Multi-Source (MS) refers to the MSDA methods. We report the multi-run statistics (mean and standard deviation) obtained over three different runs.

\textbf{c) Implementation Details.} We implement our approach in PyTorch~\cite{NEURIPS2019_9015_pytorch}. We use the Adam~\cite{adam} optimizer, with learning rate $10^{-5}$ and weight decay $5 \times 10^{-4}$ for stochastic optimization.
The losses in Eq.~\ref{eq:our_mirror_loss} and Eq.~\ref{eq:our_target_mirror_loss} are alternatively optimized and the target agreement rate (Eq.~\ref{eq:agreement_rate}) is periodically monitored for convergence. We set $n_e = 15~\text{epochs}$ as the update rate for the target pseudo-labels (line {\color{red}12} in Algo.~{\color{red}1}). The total number of training iterations are decided based on the convergence of the target agreement rate $A$ for each dataset. Following prior MSDA approaches~\cite{zhu2019aligning_mfsan,peng2019moment}, we use ResNet-50 (SImpAl$_{50}$) and ResNet-101 (SImpAl$_{101}$)~\cite{he2016deep_resnet} as the CNN backbone. 

\newcommand{\CC}{\cellcolor{gray!14}{}}

\begin{table}[!t]
	
	\caption{\textbf{Results on five standard benchmark datasets.} `SB' stands for Single Best, `SC' stands for Source Combined, and `MS' denotes MSDA methods. The results for prior baselines are reported from \cite{peng2019moment} and \cite{zhu2019aligning_mfsan}. See Supplementary for the full comparison table on DomainNet.}
    \label{table:all_data_results}
    \vspace{1mm}

\center

\resizebox{1\linewidth}{!}{

\begin{tabular}{ cc }   

\begin{tabular}{c} 
A. \textbf{\textit{Office-31}} \\ 

	 \adjustbox{valign=t}{
		\setlength\tabcolsep{6.8pt} 
		\begin{tabular}{|c|c|ccc|c|}
		    \hline
			&\textbf{Method}
			&$\rightarrow$\textbf{D}
			&$\rightarrow$\textbf{W} 
			&$\rightarrow$\textbf{A} 
			&\textbf{Avg}  \\
			
			
			\hline
			\multirow{6}{*}{\begin{turn}{90}\makecell{SB}\end{turn}} &Source Only&99.3&96.7&62.5&86.2\\
                &DDC&98.2&95.0&67.4&86.9\\
                &DAN&99.5&96.8&66.7&87.7\\
                &D-CORAL&99.7&98.0&65.3&87.7\\
                &RevGrad&99.1&96.9&68.2&88.1\\
                &RTN&99.4&96.8&66.2&87.5\\
			
			\hline
			\multirow{3}{*}{\begin{turn}{90}\makecell{SC}\end{turn}} &DAN&99.6&97.8&67.6&88.3\\
                &D-CORAL&99.3&98.0&67.1&88.1\\
                &RevGrad&99.7&98.1&67.6&88.5\\
			
			\hline	\multirow{4}{*}{\begin{turn}{90}\makecell{MS}\end{turn}}&DCTN&99.3& 98.2& 64.2&87.2\\
                    &MFSAN&99.5& 98.5 & 72.7 & 90.2 \\
                    \cline{2-6}
                    \cline{2-6}
                &\CC SImpAl$_{50}$ 
                & \CC 99.2$^{\pm\text{0.2}}$ 
                & \CC 97.4$^{\pm\text{0.1}}$ 
                & \CC 70.6$^{\pm\text{0.6}}$ 
                & \CC 89.0$^{\pm\text{0.3}}$\\
                &\CC SImpAl$_{101}$
                &\CC 99.4$^{\pm\text{0.2}}$
                &\CC 97.9$^{\pm\text{0.2}}$
                &\CC 71.2$^{\pm\text{0.4}}$
                &\CC 89.5$^{\pm\text{0.3}}$\\
            \hline
        
		\end{tabular}}\\
		\vspace{-1mm}
		\\
		C. \textbf{\textit{Office-Caltech}} \\ 
		\adjustbox{valign=t}{
	
	\setlength\tabcolsep{3.0pt} 
	
	\begin{tabular}{|c|c|cccc|c|}
			\hline
			&\textbf{Method}
			&$\rightarrow$\textbf{W}
			&$\rightarrow$\textbf{D}
			&$\rightarrow$\textbf{C}
			&$\rightarrow$\textbf{A}
			&\textbf{Avg}
			\\ 
			\hline
			\multirow{2}{*}{\begin{turn}{90}\makecell{SC}\end{turn}}	&Source Only& 99.0	& 98.3& 87.8	& 86.1	  &	92.8	\\
			&DAN		& 99.3	& 98.2 & 89.7	& {
			94.8}	  &	95.5	\\
			\hline				
			\multirow{9}{*}{\begin{turn}{90}\makecell{MS}\end{turn}}
			&Source Only & 99.1  & 98.2 & 85.4 &88.7 & 92.9 \\				
			&DAN	&99.5	&	99.1	& 89.2	& 91.6 	& 94.8	\\
			&DCTN & 99.4 & 99.0 & 90.2 & 92.7 & 95.3\\
			&JAN	&99.4	&	99.4	& 91.2	& 91.8 	& 95.5\\
			&MEDA & 99.3 & 99.2 & 91.4 & 92.9 & 95.7 \\
			&MCD & 99.5 & 99.1 & 91.5 & 92.1 & 95.6 \\
			
			&M\textsuperscript{3}SDA	&99.4		&99.2	& 91.5	&94.1 &	96.1	\\
        \cline{2-7}
        \cline{2-7}
        & \CC SImpAl$_{50}$
        & \CC 99.3$^{\pm\text{0.1}}$
        & \CC 99.8$^{\pm\text{0.1}}$
        & \CC 92.2$^{\pm\text{0.1}}$
        & \CC 95.3$^{\pm\text{0.2}}$
        & \CC 96.7$^{\pm\text{0.1}}$\\
        & \CC SImpAl$_{101}$
        & \CC 100$^{\pm\text{0.0}}$
        & \CC 100$^{\pm\text{0.0}}$
        & \CC 94.6$^{\pm\text{0.2}}$
        & \CC 95.6$^{\pm\text{0.3}}$
        & \CC 97.5$^{\pm\text{0.1}}$\\
        
        \hline
        
		\end{tabular}}
		\end{tabular}
		
		&
		\begin{tabular}{c} 
		B. \textbf{\textit{ImageCLEF}} \\ 
		\adjustbox{valign=t}{
		\begin{tabular}{|c|c|ccc|c|}
			\hline
			&\textbf{Method}
			&$\rightarrow$\textbf{P}
			&$\rightarrow$\textbf{C}
			&$\rightarrow$\textbf{I}
			&\textbf{Avg}  \\
			
			\hline
			\multirow{6}{*}{\begin{turn}{90}\makecell{SB}\end{turn}} &Source Only&74.8&91.5&83.9&83.4\\
			&DDC&74.6&91.1&85.7&83.8\\
			&DAN&75.0&93.3&86.2&84.8\\
			&D-CORAL&76.9&93.6&88.5&86.3\\
			&RevGrad&75.0& 96.2 &87.0&86.1\\
			&RTN&75.6&95.3&86.9&85.9\\
			
			\hline
			\multirow{3}{*}{\begin{turn}{90}\makecell{SC}\end{turn}} &DAN&77.6&93.3&92.2&87.7\\
			&D-CORAL&77.1&93.6&91.7&87.5\\
			&RevGrad&77.9&93.7&91.8&87.8\\
			
			\hline	\multirow{4}{*}{\begin{turn}{90}\makecell{MS}\end{turn}}
			&DCTN&75.0&95.7&90.3&87.0\\
            &MFSAN&79.1&95.4&93.6&89.4\\
        \cline{2-6}
        \cline{2-6}    
        &\CC SImpAl$_{50}$
        &\CC 77.5$^{\pm\text{0.3}}$
        &\CC 93.3$^{\pm\text{0.3}}$
        &\CC 91.0$^{\pm\text{0.4}}$
        &\CC 87.3$^{\pm\text{0.3}}$\\
        &\CC SImpAl$_{101}$
        &\CC 78.0$^{\pm\text{0.5}}$
        &\CC 95.2$^{\pm\text{0.5}}$
        &\CC 91.7$^{\pm\text{0.4}}$
        &\CC 88.3$^{\pm\text{0.5}}$\\
        
        \hline
        
		\end{tabular}}
		\\
		\vspace{-1mm}
		\\
		D. \textbf{\textit{Office-Home}} \\
		\adjustbox{valign=t}{
		\setlength\tabcolsep{3.08pt} 
		\begin{tabular}{|c|c|cccc|c|}
		    \hline
			&\textbf{Method}
			&$\rightarrow$\textbf{Ar}
			&$\rightarrow$\textbf{Cl}
			&$\rightarrow$\textbf{Pr}
		    &$\rightarrow$\textbf{Rw}	
		    &\textbf{Avg}  \\

			\hline
			\multirow{5}{*}{\begin{turn}{90}SB \end{turn}}&Source Only&65.3&49.6&79.7&75.4&67.5\\
            &DDC&64.1&50.8&78.2&75.0&67.0\\
            &DAN&68.2&56.5&80.3&75.9&70.2\\
            &D-CORAL&67.0&53.6&80.3&76.3&69.3\\
            &RevGrad&67.9&55.9& 80.4 &75.8&70.0\\
			
			\hline
			\multirow{3}{*}{\begin{turn}{90}SC \end{turn}} &DAN&68.5&59.4&79.0& 82.5&72.4\\
&D-CORAL&68.1&58.6&79.5&82.7&72.2\\
&RevGrad&68.4&59.1&79.5&82.7&72.4\\

			\hline	\multirow{3}{*}{\begin{turn}{90}MS \end{turn}} 
            &MFSAN& 72.1 &62.0&80.3&81.8&74.1\\
            \cline{2-7}
            \cline{2-7}
            &\CC SImpAl$_{50}$
            &\CC 70.8$^{\pm\text{0.2}}$
            &\CC 56.3$^{\pm\text{0.2}}$
            &\CC 80.2$^{\pm\text{0.3}}$
            &\CC 81.5$^{\pm\text{0.3}}$
            &\CC 72.2$^{\pm\text{0.6}}$\\
            &\CC SImpAl$_{101}$
            &\CC 73.4$^{\pm\text{0.4}}$
            &\CC 62.4$^{\pm\text{0.1}}$
            &\CC 81.0$^{\pm\text{0.2}}$
            &\CC 82.7$^{\pm\text{0.2}}$ 
            &\CC 74.8$^{\pm\text{0.2}}$\\
        
            \hline
		\end{tabular}}
		
		\end{tabular}
		\end{tabular}
} \\
\vspace{1mm}
\begin{tabular}{cc}
\multicolumn{2}{c}{
{\footnotesize E. \textbf{\textit{DomainNet}}} 
\vspace{-1mm}
}
\\
\multicolumn{2}{c}{
	\resizebox{0.8\textwidth}{!}{
	 \adjustbox{valign=t}{
		\setlength\tabcolsep{6.8pt} 
		\begin{tabular}{|c|c|cccccc|c|}
		    \hline
			&\textbf{Method}
			&$\rightarrow$\textbf{Clp}
			&$\rightarrow$\textbf{Inf} 
			&$\rightarrow$\textbf{Pnt} 
			&$\rightarrow$\textbf{Qdr} 
			&$\rightarrow$\textbf{Rel} 
			&$\rightarrow$\textbf{Skt} 
			&\textbf{Avg}  \\
			
			
			\hline	\multirow{2}{*}{\begin{turn}{90}\makecell{MS}\end{turn}}&M$^3$SDA 
			&57.2$^{\pm\text{0.9}}$ 
			&24.2$^{\pm\text{1.2}}$ 
			&51.6$^{\pm\text{0.4}}$ 
			&5.2$^{\pm\text{0.4}}$ 
			&61.6$^{\pm\text{0.9}}$ 
			&49.6$^{\pm\text{0.5}}$
			&41.5$^{\pm\text{0.7}}$\\
                    \cline{2-9}
                    \cline{2-9}
            &\CC SImpAl$_{101}$
            &\CC 66.4$^{\pm\text{0.8}}$
            &\CC 26.5$^{\pm\text{0.5}}$
            &\CC 56.6$^{\pm\text{0.7}}$
            &\CC 18.9$^{\pm\text{0.8}}$
            &\CC 68.0$^{\pm\text{0.5}}$
            &\CC 55.5$^{\pm\text{0.3}}$
            &\CC 48.6$^{\pm\text{0.6}}$\\
            \hline
        
		\end{tabular}}
	}
}
\end{tabular}

\vspace{-4mm}
\end{table}

\subsection{Results}

We present the results in Table~\navref{table:all_data_results}. The results for the prior baselines are reported from \cite{peng2019moment} and \cite{zhu2019aligning_mfsan}. Due to the limits of space, we present the full comparison table for \textit{DomainNet} in the Supplementary. 


\textit{Office-31}~\cite{office} dataset has 4652 images across Amazon (\textbf{A}), DSLR (\textbf{D}) and Webcam (\textbf{W}) domains having 31 object classes found in an office environment. 
\textit{ImageCLEF}\footnote{\url{http://imageclef.org/2014/adaptation.}} dataset has been created by selecting 12 shared classes among \textit{ImageNet} (\textbf{I}) \cite{imagenet}, \textit{Caltech-256} (\textbf{C}) \cite{griffin2007caltech}, \textit{Pascal-VOC 2012} (\textbf{P}) \cite{Everingham10Pascal_voc}, with 600 images per domain. 
\textit{Office-Caltech} \cite{gong2012geodesic} dataset consists of 2533 images across 10 classes shared between \textit{Caltech-256} (\textbf{C}) and the three domains of \textit{Office-31} (\textbf{A}, \textbf{D}, \textbf{W}). 
\textit{Office-Home} \cite{venkateswara2017deep} is a more challenging medium-scale dataset containing about 15588 images in 4 domains: Art (\textbf{Ar}),
Clipart (\textbf{Cl}), Product (\textbf{Pr}) and Real-World (\textbf{Rw}), sharing 65 categories of objects found in the office and home environments.
\textit{DomainNet} \cite{peng2019moment} dataset is the largest and the most challenging benchmark, containing 6 diverse domains, with 345 classes, and around 0.6 million images. 
\subsection{Analysis}
\label{subsec:analysis}

\begin{figure}[!t]
    \centering
    \includegraphics[width=1\linewidth]{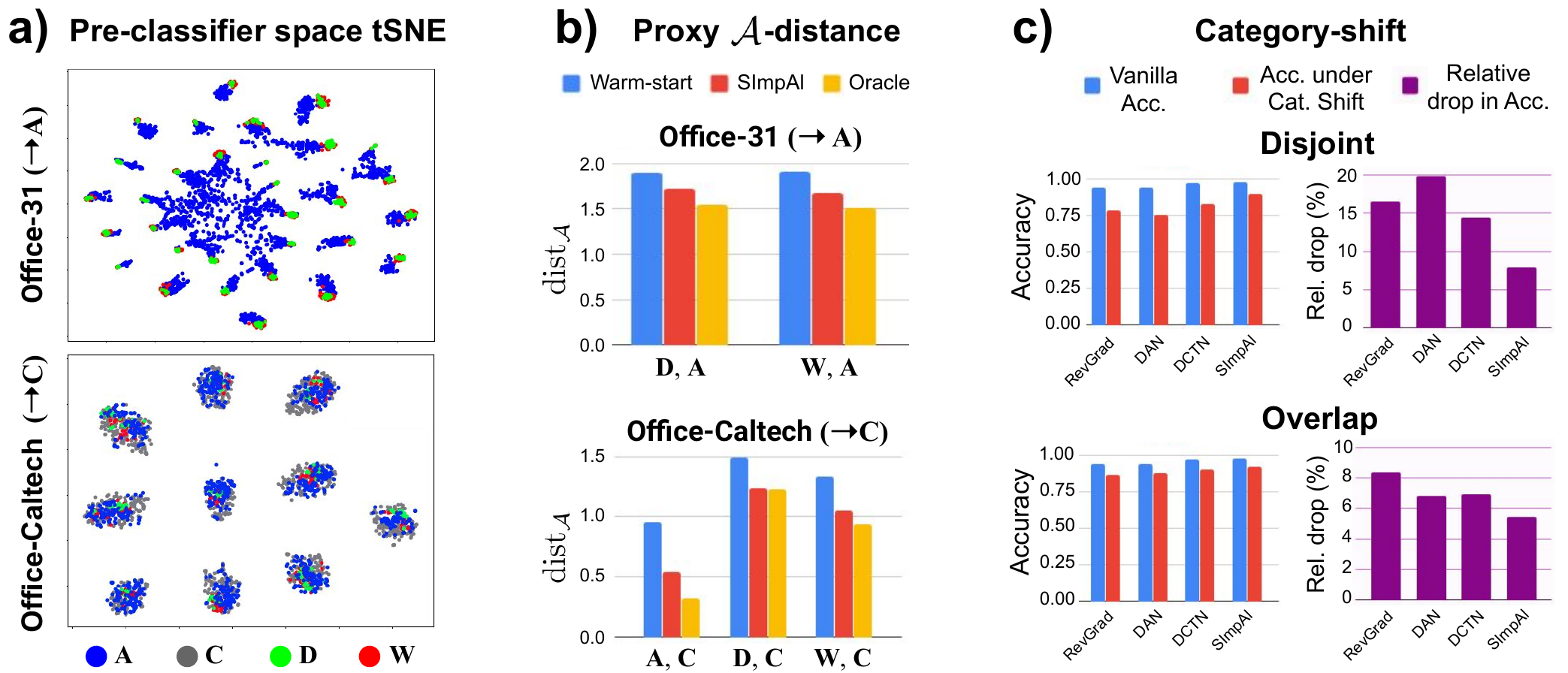}
    \vspace{-5mm}
    \caption{\small \textbf{(a) t-SNE.} We show alignment of domains at the pre-classifier space ($f$-output). \textbf{(b) Proxy $\mathcal{A}$-distance ($\downarrow$).} The values of $\operatorname{dist}_{\mathcal{A}}$ (Sec. \navref{subsec:analysis}{a}) are obtained between each source-target pair for the corresponding models shown in (a). 
    \textbf{(c) Performance drop under category-shift ($\downarrow$).} Following \cite{dctn}, we compare SImpAl against RevGrad~\cite{NormalAdaptfirstBackprop}, DAN~\cite{long2015learning}, DCTN~\cite{dctn} in the Overlap and Disjoint scenarios on Office-31 (\textbf{A}, \textbf{D} $\rightarrow$ \textbf{W}).
    }
    \label{fig:tsne_pad_ssm_cat_shift}
    \vspace{-2mm}
\end{figure}

\textbf{a) Implicit alignment of features.} In Fig.~\navref{fig:tsne_pad_ssm_cat_shift}{a}, we plot the t-SNE~\cite{tsne} embeddings of the features at the pre-classifier space (output of $f$) for SImpAl. Further, we calculate the Proxy-$\mathcal{A}$ distance~\cite{ben2010theory} defined as $\operatorname{dist}_{\mathcal{A}} = 2(1-2\epsilon)$ where $\epsilon$ is the generalization error of a domain discriminator. In Fig.~\navref{fig:tsne_pad_ssm_cat_shift}{b}, we report the $\operatorname{dist}_{\mathcal{A}}$ value across each source-target pair for 3 different models - 1) warm-start model, trained on the source domains, 2) the model after adaptation using SImpAl, 3) an oracle model employing SImpAl, where the target pseudo-labels are replaced by the ground-truth labels. This shows that adaptation using SImpAl effectively reduces the distribution-shift in the latent space. Further, we also demonstrate implicit alignment under large domain-shifts (such as Quickdraw and Real-world domains on DomainNet), which enables applications such as cross-domain image retrieval on an unlabeled target domain. See Suppl. for further analysis on implicit alignment.

\begin{figure}
    \centering
    \includegraphics[width=1\linewidth]{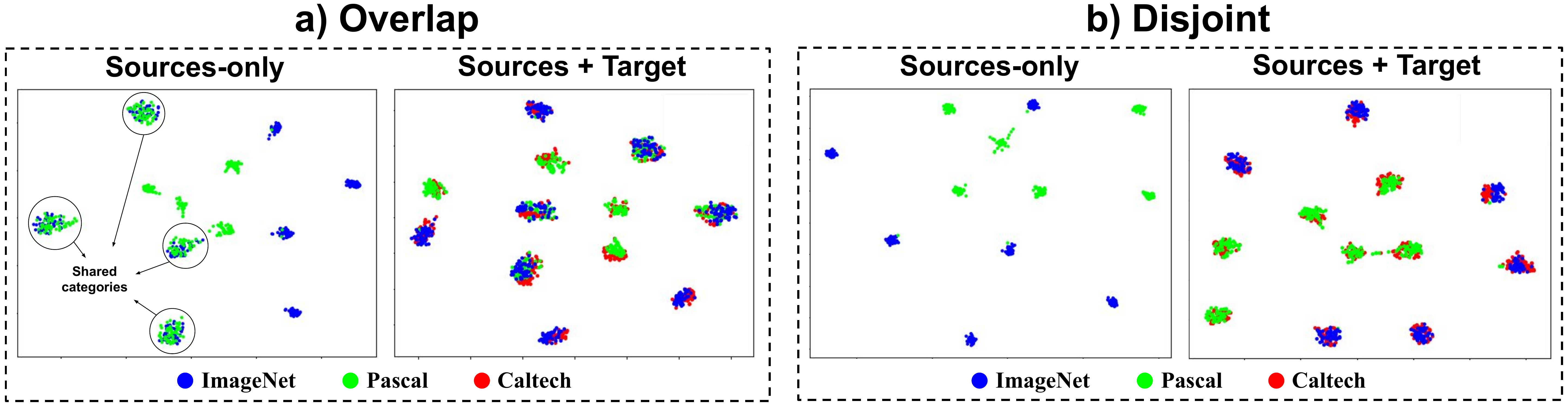}
    \vspace{-5mm}
    \caption{\small Pre-classifier space t-SNE embeddings for category-shift scenarios on ImageCLEF (\textbf{I}, \textbf{P} $\rightarrow$ \textbf{C}). Two plots are shown: features of the source domains only, and, features of all the domains. Best viewed in color. \textbf{(a) Overlap.} Here, 4 classes are shared between the sources ($|\mathcal{C}_{s_1} \cap \mathcal{C}_{s_2}| = 4$). Observe that the shared classes align, while the domain-specific classes are clustered separately. \textbf{(b) Disjoint.} In the absence of shared classes ($\mathcal{C}_{s_1} \cap \mathcal{C}_{s_2} = \phi$), class-wise alignment is not observed among the sources, which is essential to avoid negative-transfer. Note, in both cases, the target clusters align with the respective source clusters (since $\mathcal{C} = \mathcal{C}_{s_1} \cup \mathcal{C}_{s_2}$). This demonstrates implicit alignment under category-shift. See Suppl. for wider trends.}
    \label{fig:tsne_catshift_figure}
    \vspace{-2mm}
\end{figure}

\textbf{b) Extension to category-shift.} To present a more practical scenario for MSDA, \cite{dctn} introduced two category-shift settings - overlap and disjoint, where the source domains contain overlapping label sets (\ie $\mathcal{C}_{s_i} \cap \mathcal{C}_{s_{i'}} \neq \phi$, but $\mathcal{C}_{s_i} \cap \mathcal{C}_{s_{i'}} \neq \mathcal{C}_{s_i} \cup \mathcal{C}_{s_{i'}}$) and disjoint label-sets ($\mathcal{C}_{s_i} \cap \mathcal{C}_{s_{i'}} = \phi$) respectively. In such scenarios, it is vital to prevent mis-alignment of different classes across the source domains to avoid negative transfer \cite{mada}. Furthermore, since prior MSDA approaches learn domain-specific classifiers, they require separate mechanisms to obtain class probabilities for the domain-specific and the shared classes separately~\cite{dctn}. However, our approach remains unmodified under the presence of category-shift; as such, each classifier learns all the target classes, and the computation of the class probabilities (Eq.~\ref{eq:ensemble_prediction}) remains unchanged. Fig. \navref{fig:tsne_pad_ssm_cat_shift}{c} shows that category-shift is a challenging scenario where all methods show performance degradation, however SImpAl is found to exhibit a relatively lower degradation in the target performance. This is supported by the observation that even under category-shift, only the shared classes align as shown in Fig.~\ref{fig:tsne_catshift_figure}. See Suppl. for further analysis.

\begin{figure}[!t]  
    \centering
    \includegraphics[width=1\linewidth]{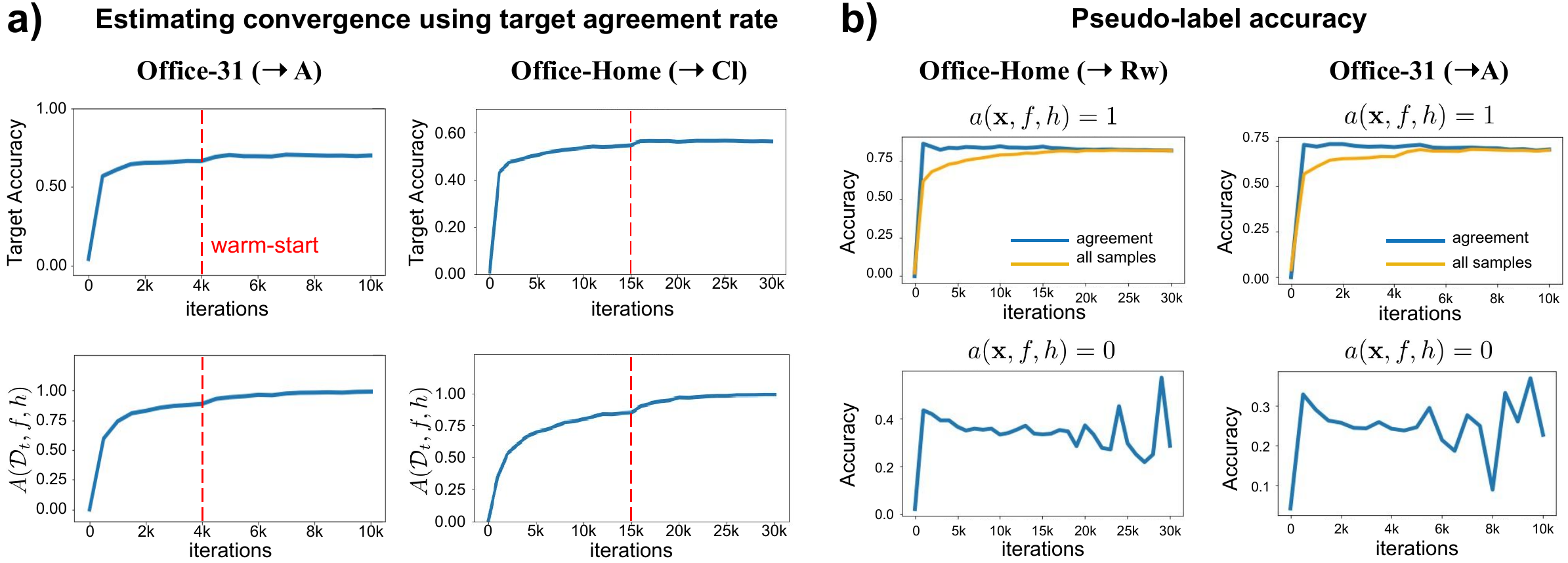}
    \vspace{-4mm}
    \caption{\small \textbf{(a) Target agreement rate $A(\mathcal{D}_t, f, h)$.} The target agreement rate (bottom) can be used as a cue to determine training convergence (top). \textbf{(b) Pseudo-label accuracy.} We observe a higher pseudo-label accuracy for target samples with $a(\mathbf{x}_t, f, h) = 1$ (top, blue curve), as compared to those with $a(\mathbf{x}, f, h) = 0$ (bottom). 
    }
    \label{fig:target_agreement_rate}
\end{figure}

\textbf{c) Target Agreement Rate.} Fig.~\navref{fig:target_agreement_rate}{a} shows the trend in the target agreement rate ($A(\mathcal{D}_t, f, h)$) and target performance as training proceeds. We make two observations. Firstly, we find that $A$ increases during training, indicating that the target samples migrate into the classifier agreement region in the latent space ($\{f(\mathbf{x}_t) ~|~ a(\mathbf{x}_t, f, h) = 1\}$). This migration is necessary for a successful adaptation since the source domains inherently fall in the classifier agreement region (due to the nature of the source training for warm-start). Secondly, a correspondence between the convergence of the target agreement rate and the target accuracy is seen, which validates our hypothesis that $A$ can be used as a cue to determine the training convergence. This result is of interest in Unsupervised Domain Adaptation methods where the requirement of target labels has been the de-facto for model selection.

\begin{figure}[!t]
    \centering
    \includegraphics[width=1\linewidth]{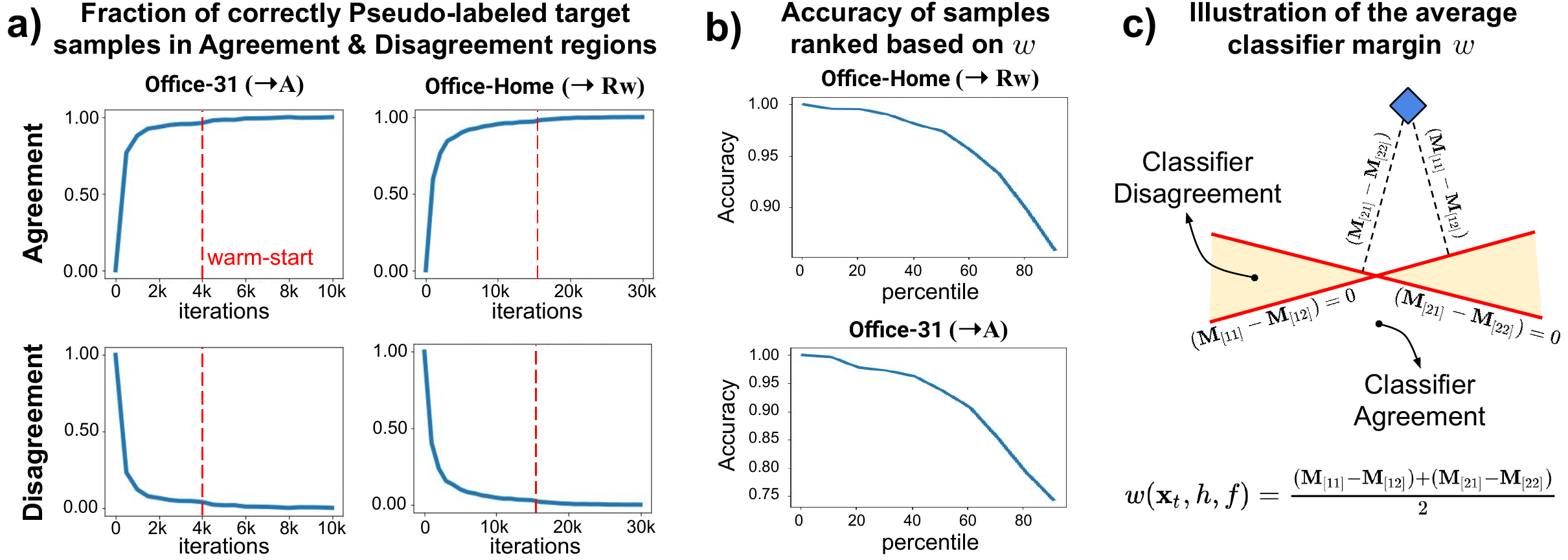}
    \vspace{-4mm}
    \caption{\small \textbf{(a) Migration of target samples with correct pseudo-labels.} During training, we find that the fraction of target samples with correct pseudo-label predictions increases in the agreement region. \textbf{(b) Curriculum using $w(\mathbf{x}_t, f, h)$.} When sorted in descending order based on $w$, the target samples exhibit an easy-to-hard curriculum. \textbf{(c) Geometrical interpretation of $w$.} The figure shows the significance of $w$ in a two-source two-class scenario (as in Fig.~\ref{fig:basic_overview}). Intuitively, $w$ is a measure of classifier confidence averaged over each classifier. Target samples that are further into the classifier agreement region exhibit a higher value of $w$.}
    \label{fig:target_migration}
\end{figure}

\textbf{d) Do the classifiers agree on correct pseudo-labels?} We also calculate the classifier agreement (and disagreement) for target samples that are pseudo-labeled correctly. Notably, Fig.~\navref{fig:target_migration}{a} demonstrates that the classifiers tend to agree on an increasing number of target samples with correct pseudo-label predictions.
This motivates the periodic update of ${\mathcal{D}_t}'$ (Lines {\color{red} 12-13} in Algo. {\color{red} 1}), which captures an increasing number of target samples with correct pseudo-labels, as the adaptation proceeds.

\textbf{e) How accurate are target pseudo-labels?} As described in Sec.~\ref{subsec:target_tuning}, we use classifier agreement to select target samples (${\mathcal{D}_t}'$) with a higher pseudo-label accuracy. In Fig. \navref{fig:target_agreement_rate}{b}, we plot the accuracy of pseudo-labels separately for target samples having classifier agreement (\ie $a(\mathbf{x}_t, f, h) = 1$) and disagreement (\ie $a(\mathbf{x}_t, f, h) = 0$). Clearly, pseudo-labels are more accurate (more reliable) when the classifiers agree. Further, the accuracy on the target samples with agreement, ${\mathcal{D}_t}'$, is higher than the accuracy on all target samples, $\mathcal{D}_t$ (orange curve in Fig.~\navref{fig:target_agreement_rate}{b}). Thus, the use of ${\mathcal{D}_t}'$ with a higher accuracy in pseudo-labels plays a key role in gradually improving the target performance.

\textbf{f) Using curriculum for target samples.} We form a curriculum for the target samples using the average classifier margin $w(\mathbf{x}_t, f, h)$ as a weight. Fig. \navref{fig:target_migration}{c} shows the geometrical interpretation of $w$, that measures how far into the agreement region a target sample falls. Thus, $w$ can be seen as a measure of the confidence in the prediction. As studied by prior methods \cite{mixtureofexperts,ruder2017knowledgeadaptation,sohn2020fixmatch}, high confidence predictions are often correct. We show this in Fig.~\navref{fig:target_migration}{b} where we plot the precision of target pseudo-labels at various confidence percentiles (in descending order of $w$). The accuracy shows a decreasing trend with $w$, validating our hypothesis that $w$ yields an easy-to-hard curriculum. Although our framework supports confidence thresholding to further minimize the pseudo-label noise, we do not employ thresholds for the main results (Table~\ref{table:all_data_results}) as it introduces sensitive hyperparameters. See Supplementary for an empirical analysis with confidence thresholding.









\section{Conclusion}

In this paper, we demonstrated Self-supervised Implicit Alignment (SImpAl), that serves as a simple method to perform Multi-Source Domain Adaptation (MSDA). We observed that deep models exhibit the potential to implicitly align features under label supervision, even in the presence of domain-shift. We demonstrated the use of classifier agreement in SImpAl - to obtain pseudo-labeled target samples, to perform latent space alignment and to determine the training convergence. Extensive empirical analysis demonstrates the efficacy of SImpAl for MSDA. 

Our work can facilitate the study of simple and effective algorithms for unsupervised domain adaptation. The insights obtained from our study can be used to explain the efficacy of a number of related self-supervised approaches. A potential direction of research is to develop efficient adaptation algorithms that are devoid of sensitive hyperparameters. Exploring SImpAl for scenarios such as Universal Domain Adaptation~\cite{UDA_2019_CVPR} would also be of future interest.






\section*{Broader Impact}

This work presents a simple and effective solution for Multi-Source Domain Adaptation, that has a two-fold positive impact. First, the method is aimed at improving the performance of prediction models by mitigating the bias caused by domain-shift between the training dataset and the test data encountered when deployed in a real-world environment. This is of growing interest in the machine learning community. Secondly, the insights presented in this work facilitate the study of efficient methods to perform domain adaptation, motivating the innovation of, for instance, energy-efficient methods to generalize deep models. While the method shows promising results under domain-shift, one should be cautious of the use of the pseudo-labeling procedure in the presence of adversarial samples, where the pseudo-labels may be less reliable and may result in performance degradation.

\begin{ack}

This work was supported by a project grant from MeitY (No.4(16)/2019-ITEA), Govt. of India and a WIRIN project. We would also like to thank the anonymous reviewers for their valuable suggestions.

\end{ack}



\section*{Supplementary material}

# Supplementary: Your Classifier can Secretly Suffice Multi-Source Domain Adaptation

| Section | Content |
|---|---|
| Sec. 1 | Results on *DomainNet* |
| Sec. 2 | Further analysis on Implicit Alignment |
| Sec. 2.1 | - Alignment under Category-shift |
| Sec. 2.2 | - Is classifier agreement necessary? |
| Sec. 2.3 | - Ablations using thresholding schemes |
| Sec. 3 | Practical Application: Cross-Domain Image Retrieval |
| Sec. 4 | Code Reference |

## 1 Results on DomainNet

Owing to the limits of space, we present a summary of results on *DomainNet* in the paper. The full comparison is presented here in Table 1. The results for the prior arts are reported from [9].

Table 1: Results on the *DomainNet* [9] dataset.

| | Method | →Clp | →Inf | →Pnt | →Qdr | →Rel | →Skt | Avg |
|---|---|---|---|---|---|---|---|---|
| SB | Source Only | 39.6±0.5 | 8.2±0.7 | 33.9±0.6 | 11.8±0.6 | 41.6±0.8 | 23.1±0.7 | 26.4±0.7 |
| | DAN [5] | 39.1±0.5 | 11.4±0.8 | 33.3±0.6 | 16.2±0.3 | 42.1±0.7 | 29.7±0.9 | 28.6±0.6 |
| | RTN [7] | 35.3±0.7 | 10.7±0.6 | 31.7±0.8 | 13.1±0.6 | 40.6±0.5 | 26.5±0.7 | 26.3±0.7 |
| | JAN [6] | 35.3±0.7 | 9.1±0.6 | 32.5±0.6 | 14.3±0.6 | 43.1±0.7 | 25.7±0.6 | 26.7±0.6 |
| | ADDA [12] | 39.5±0.8 | 14.5±0.6 | 29.1±0.7 | 14.9±0.5 | 41.9±0.8 | 30.7±0.6 | 28.4±0.7 |
| | MCD [11] | 42.6±0.3 | 19.6±0.7 | 42.6±0.9 | 3.8±0.6 | 50.5±0.4 | 33.8±0.8 | 32.2±0.6 |
| SC | Source Only | 47.6±0.5 | 13.0±0.4 | 38.1±0.4 | 13.3±0.3 | 51.9±0.8 | 33.7±0.5 | 32.9±0.5 |
| | DAN [5] | 45.4±0.4 | 12.8±0.8 | 36.2±0.5 | 15.3±0.3 | 48.6±0.7 | 34.0±0.5 | 32.1±0.5 |
| | RTN [7] | 44.2±0.5 | 12.6±0.7 | 35.3±0.5 | 14.6±0.7 | 48.4±0.6 | 31.7±0.7 | 31.1±0.6 |
| | JAN [6] | 40.9±0.4 | 11.1±0.6 | 35.4±0.5 | 12.1±0.6 | 45.8±0.5 | 32.3±0.6 | 29.6±0.5 |
| | ADDA [12] | 47.5±0.7 | 11.4±0.6 | 36.7±0.5 | 14.7±0.5 | 49.1±0.8 | 33.5±0.4 | 32.2±0.6 |
| | MCD [11] | 54.3±0.6 | 22.1±0.7 | 45.7±0.6 | 7.6±0.4 | 58.4±0.6 | 43.5±0.5 | 38.5±0.6 |
| MS | DCTN [13] | 48.6±0.7 | 23.5±0.5 | 48.8±0.6 | 7.2±0.4 | 53.5±0.5 | 47.3±0.4 | 38.2±0.5 |
| | M$^3$SDA [9] | 57.2±0.9 | 24.2±1.2 | 51.6±0.4 | 5.2±0.4 | 61.6±0.8 | 49.6±0.5 | 41.5±0.7 |
| | M$^3$SDA-$\beta$ [9] | 58.6±0.5 | 26.0± 0.8 | 52.3±0.5 | 6.3±0.5 | 62.7±0.5 | 49.5±0.7 | 42.6±0.6 |
| | SImpAl$_{101}$ | 66.4±0.8 | 26.5±0.5 | 56.6±0.7 | 18.9±0.8 | 68.0±0.5 | 55.5±0.3 | 48.6±0.6 |

## 2 Further Analysis on Implicit Alignment

In this section present further analysis on implicit alignment. First, we show wider trends for implicit alignment under category-shift. Then, we present ablations on the training objective (with and without classifier agreement, single classifier head). Finally, we study thresholding schemes.

### 2.1 Implicit Alignment under Category-shift

We find that SImpAl works well even under category-shift. In Fig. 4c of the paper, we compare the relative drop (%) in accuracy under category-shift which is obtained as,

$$\text{relative drop} = \frac{(A_{vanilla} - A_{category-shift})}{A_{vanilla}} \times 100 \qquad (1)$$

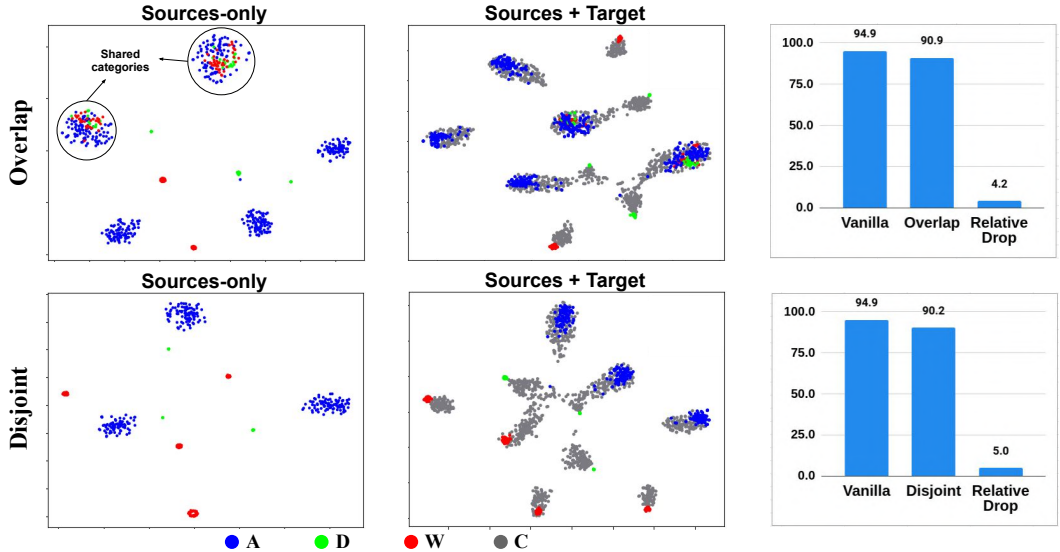

Figure 1: Pre-classifier space ($f$-output) t-SNE showing the alignment of features under the category-shift settings **Overlap** (top) and **Disjoint** (bottom) on the *Office-Caltech* dataset for the task **A, D, W**→**C**. Note the alignment of selected clusters (shared classes among the source domains) in the **Overlap** scenario, while none of the source clusters align in the **Disjoint** scenario. Best viewed in color.

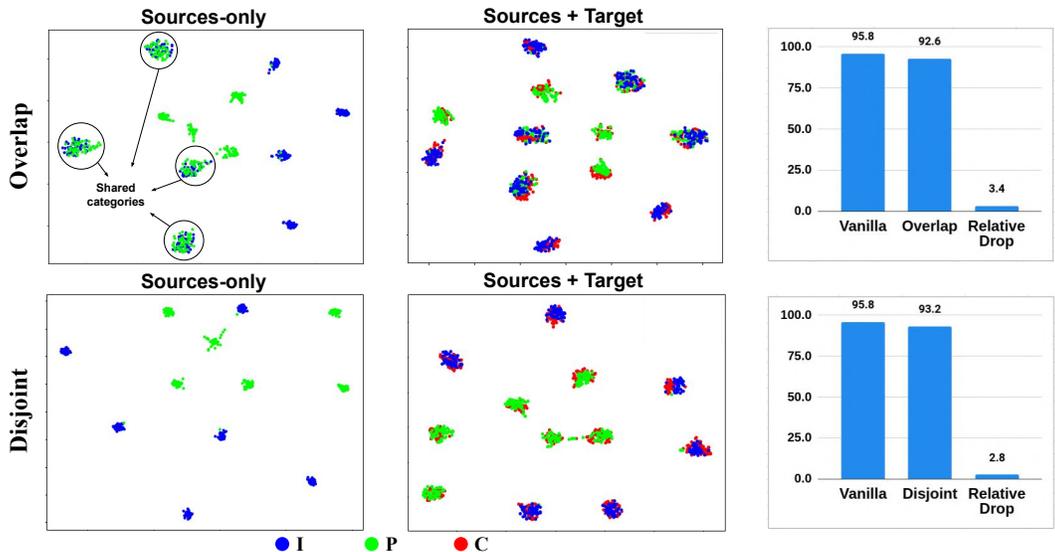

Figure 2: Pre-classifier space t-SNE showing the alignment of features under the category-shift settings **Overlap** (top) and **Disjoint** (bottom) on the *ImageCLEF* dataset for the task **I, P**→**C**. Best viewed in color.

Our approach exhibits a relatively lower drop in accuracy. To understand this better, we further investigate the latent space alignment of the domains after adaptation using SImpAl. We show two t-SNE plots, 1) Sources-only, *i.e.* showing alignment among the source domains, and 2) Sources + Target, *i.e.* showing the alignment among all the domains, corresponding to an adapted model.

Fig. 1 shows the latent space alignment for the task **A, D, W**→**C** for the *Office-Caltech* dataset. For the **Overlap** scenario, we set the number of shared categories to 2, while the number of source-private categories are 3, 3, 2 for the sources Amazon (**A**), DSLR (**D**) and Webcam (**W**) respectively. Clearly, alignment among the sources is observed only for the two shared categories (annotated in Fig. 1). In contrast, for the **Disjoint** scenario, the source domains **A**, **D**, **W** contain 3, 3, 4 unique classes respectively. Here, none of the source clusters align as expected (since each source has a distinct set of classes). However, in both scenarios, we find that the target domain aligns with at least one source

2

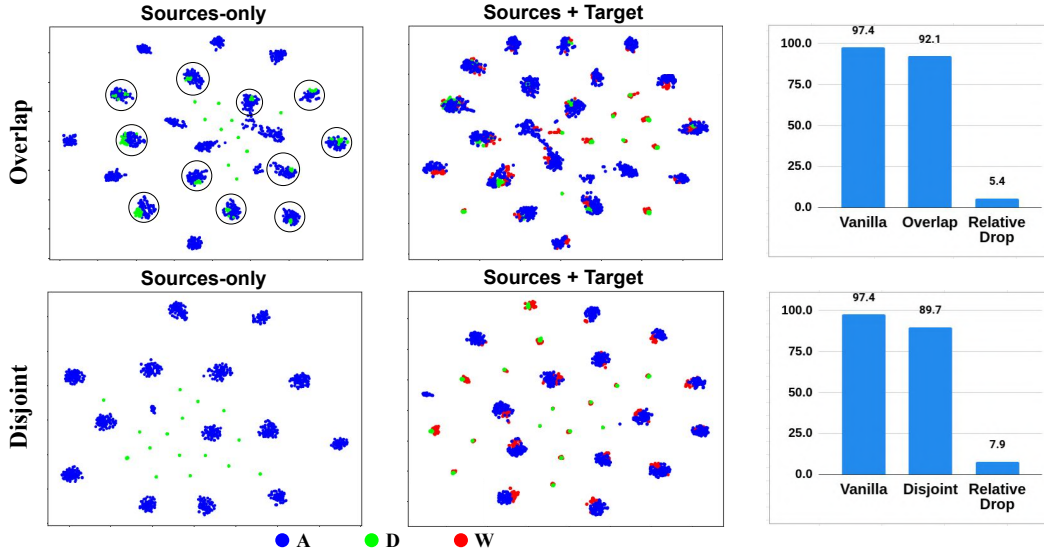

Figure 3: Pre-classifier space t-SNE showing the alignment of features under the category-shift settings **Overlap** (top) and **Disjoint** (bottom) on the *Office-31* dataset for the task **A, D→W** (in line with Fig. 4c of the paper). Source clusters corresponding to the shared classes are encircled. Best viewed in color.

domain (multiple sources if categories are shared). We find a similar trend across other datasets as shown in Fig. 2 for the **I, P→C** task of *ImageCLEF* (**Overlap**: 4 shared and 4, 4 source-private classes; **Disjoint**: 6, 6 source-private classes), and in Fig. 3 for the **A, D→W** task of *Office-31* (**Overlap**: 11 shared and 10, 10 source-private classes; **Disjoint**: 16, 15 source-private classes).

These results support the observation that implicit alignment can be leveraged even under category-shift. Note that, in methods such as moment matching [9] or adversarial alignment [2], there is usually no enforcement of class-level alignment across the domains. Thus, such methods are prone to negative transfer via conditional mis-alignment [4, 8], *i.e.* alignment of dissimilar classes. However, SImpAl enables class-conditional alignment by virtue of the label supervision, which could explain the relatively lower drop in performance under category-shift.

### 2.2 Is classifier agreement necessary?

We now turn towards the key observation that we present in the paper. By enforcing classifier agreement we are able to perform adaptation, while, when we learn domain-specific classifiers following previous MSDA approaches [13, 14], the latent space alignment is not seen. Here, we provide further empirical analysis in support of the observation through ablations on SImpAl$_{50}$.

**a) Warm-start with domain-specific classifiers.** We modify the loss formulation in Eq. 3 of the paper, to train domain-specific classifiers, as follows:

$$\min_{f,h} \mathbb{E}_{\mathcal{D} \in \{\mathcal{D}_{s_{i'}}\}_{i'=1}^{n_d}} \mathbb{E}_{(\mathbf{x}_{s_i}^{c_j}, y_{s_i}^{c_j}) \in \mathcal{D}} -\log(\sigma(\mathbf{M}_{[i \cdot]})_{[j]}) \qquad (2)$$

where $\sigma$ denotes the softmax activation function. Essentially, an instance $\mathbf{x}_{s_i}^{c_j}$ pertaining to the source domain $\mathcal{D}_{s_i}$ trains the corresponding classifier head (note that the logarithmic term contains the probability of class $c_j$ of the classifier corresponding to $s_i$). This is in line with the prior methods such as [3, 13] where the domain-specific classifiers progressively learn to discriminate among the classes in their respective domains. Thus, there is no explicit enforcement of classifier agreement.

We perform warm-start with the loss formulation in Eq. 2 (name this model, "w/o agreement") and compare against another model learned using Eq. 3 of the paper (name this model, "with agreement"). Both models are trained for the same number of iterations under identical conditions. Finally, we test each model's performance on the target domain at warm-start. The multi-run statistics over 3 random seeds are presented in Table 2. The model "with agreement" shows a consistent improvement in performance in each scenario, which suggests that the model generalizes better to the target domain.

3

Table 2: Target accuracy of warm-start models with ablation on the learning approach - "w/o agreement" (learning domain-specific classifiers) vs. "with agreement" (our approach). Refer to Sec. 2.2a for discussion.

| Model | Office-31 ($\rightarrow$A) | ImageCLEF ($\rightarrow$P) | Office-Caltech ($\rightarrow$C) | Office-Home ($\rightarrow$Ar) |
|---|---|---|---|---|
| w/o agreement | 65.8 $\pm$ 0.3 | 75.8 $\pm$ 0.3 | 89.2 $\pm$ 0.3 | 67.3 $\pm$ 0.7 |
| with agreement | 66.2 $\pm$ 0.3 | 77.0 $\pm$ 0.6 | 90.3 $\pm$ 0.4 | 68.5 $\pm$ 0.5 |

Table 3: Proxy $\mathcal{A}$-distance ($\downarrow$) measured between each pair of domains for the warm-start models "with agreement", "w/o agreement" and "single classifier". Note that the model "with agreement" consistently exhibits lower $\text{dist}_\mathcal{A}$ than "with agreement", suggesting that it aligns the domains to a greater extent (Sec. 2.2a). Furthermore, we perform an ablation by using a single classifier head which also shows lower $\text{dist}_\mathcal{A}$ (Sec. 2.2b).

| Model | Office-31 ($\rightarrow$A) | | | ImageCLEF ($\rightarrow$ P) | | |
|---|---|---|---|---|---|---|
| | $\text{dist}_\mathcal{A}$(**A, D**) | $\text{dist}_\mathcal{A}$(**A, W**) | $\text{dist}_\mathcal{A}$(**D, W**) | $\text{dist}_\mathcal{A}$(**P, C**) | $\text{dist}_\mathcal{A}$(**P, I**) | $\text{dist}_\mathcal{A}$(**I, C**) |
| w/o agreement | 1.96$\pm$0.01 | 1.96$\pm$0.01 | 1.70$\pm$0.03 | 1.25$\pm$0.03 | 0.38$\pm$0.02 | 0.94$\pm$0.02 |
| with agreement | 1.93$\pm$0.00 | 1.93$\pm$0.00 | 0.56$\pm$0.04 | 0.65$\pm$0.04 | 0.34$\pm$0.01 | 0.65$\pm$0.04 |
| single classifier | 1.91$\pm$0.03 | 1.91$\pm$0.02 | 0.52$\pm$0.2 | 1.04$\pm$0.12 | 0.38$\pm$0.06 | 0.63$\pm$0.11 |

| Model | Office-Home ($\rightarrow$ Ar) | | | | | |
|---|---|---|---|---|---|---|
| | $\text{dist}_\mathcal{A}$(**Ar, Rw**) | $\text{dist}_\mathcal{A}$(**Ar, Cl**) | $\text{dist}_\mathcal{A}$(**Ar, Pr**) | $\text{dist}_\mathcal{A}$(**Cl, Pr**) | $\text{dist}_\mathcal{A}$(**Cl, Rw**) | $\text{dist}_\mathcal{A}$(**Pr, Rw**) |
| w/o agreement | 0.70 $\pm$ 0.02 | 1.46 $\pm$ 0.05 | 1.24 $\pm$ 0.04 | 1.36 $\pm$ 0.07 | 1.38 $\pm$ 0.06 | 0.65 $\pm$ 0.00 |
| with agreement | 0.64 $\pm$ 0.01 | 1.04 $\pm$ 0.07 | 0.98 $\pm$ 0.02 | 0.69 $\pm$ 0.04 | 0.75 $\pm$ 0.02 | 0.36 $\pm$ 0.03 |
| single classifier | 0.61$\pm$0.02 | 1.16$\pm$0.10 | 1.06$\pm$0.06 | 0.90$\pm$0.15 | 0.96$\pm$0.14 | 0.44$\pm$0.06 |

| Model | Office-Caltech ($\rightarrow$ C) | | | | | |
|---|---|---|---|---|---|---|
| | $\text{dist}_\mathcal{A}$(**C, A**) | $\text{dist}_\mathcal{A}$(**C, D**) | $\text{dist}_\mathcal{A}$(**C, W**) | $\text{dist}_\mathcal{A}$(**A, D**) | $\text{dist}_\mathcal{A}$(**A, W**) | $\text{dist}_\mathcal{A}$(**D, W**) |
| w/o agreement | 1.05 $\pm$ 0.01 | 1.80 $\pm$ 0.00 | 1.78 $\pm$ 0.02 | 1.94 $\pm$ 0.01 | 1.92 $\pm$ 0.02 | 1.61 $\pm$ 0.06 |
| with agreement | 0.91 $\pm$ 0.02 | 1.49 $\pm$ 0.05 | 1.38 $\pm$ 0.04 | 1.67 $\pm$ 0.01 | 1.49 $\pm$ 0.02 | 0.92 $\pm$ 0.05 |
| single classifier | 0.97 $\pm$ 0.05 | 1.63 $\pm$ 0.09 | 1.55 $\pm$ 0.15 | 1.77 $\pm$ 0.14 | 1.66 $\pm$ 0.19 | 1.05 $\pm$ 0.18 |

To uncover the underlying effect, we measure the Proxy $\mathcal{A}$-distance [1] defined as $\text{dist}_\mathcal{A} = 2(1 - 2\epsilon)$ where $\epsilon$ is the generalization error of a domain discriminator. A lower value of this measure indicates a higher degree of alignment between the domains. In Table 3, we report $\text{dist}_\mathcal{A}$ between each pair of domains at the $f$-output space (multi-run statistics corresponding to the models trained in Table 2). Clearly, our approach exhibits a higher degree of alignment between the source domains, as compared to learning source-specific classifiers. This encourages the model to learn domain-agnostic features that are more generalizable to the target domain, resulting in an improved alignment between each source-target pair. This translates to an improvement in the target performance.

**b) Using a single classifier head.** We also train a model by replacing multiple classifiers having agreement with a single classifier. In this case, we fix the number of iterations to be the same as those obtained from SImpAl$_{50}$. In Table 4, we compare the adaptation results of the single-classifier model, against our approach (with agreement using $n_d$ classifiers). Clearly, the performance is better when using multiple classifier heads with agreement. For the single-classifier model, we find that during adaptation the performance reaches a peak and then marginally declines to a lower value before reaching the maximum number of iterations, perhaps due to noisy pseudo-labels [13]. In our approach however, classifier agreement aids in pruning those noisy samples near the decision boundaries and enhances the pseudo-labels. This is the added benefit of using multiple classifiers.

To study the extent of latent space alignment using a single classifier, we measure the Proxy $\mathcal{A}$-distance. In Table 3, we present the $\text{dist}_\mathcal{A}$ values at warm-start for the single classifier model and compare it against the two aforementioned approaches ("with" and "w/o" agreement). Particularly, we find that the $\text{dist}_\mathcal{A}$ values are either similar to the model "with agreement", or lower than the model "w/o agreement", indicating that even a single classifier enables alignment to an extent. This explains the efficacy of self-supervised approaches that use pseudo-labels for training under domain-shift.

### 2.3 Ablations using thresholding schemes

In self-training based approaches, pseudo-labeled samples are usually obtained using confidence-thresholding, *i.e.* those samples exhibiting a confidence above a certain threshold $\tau$ are chosen for self-training. However, this results in a sensitive hyperparameter $\tau$ which requires labeled target

4

Table 4: Target adaptation performance for models trained with a single classifier head ($n = 1$) and our approach with $n = n_d$ classifier heads. Refer to Sec. 2.2b for discussion.

| No. of Heads | Office-31 ($\rightarrow$A) | ImageCLEF ($\rightarrow$P) | Office-Caltech ($\rightarrow$C) | Office-Home ($\rightarrow$Ar) |
|---|---|---|---|---|
| $n = 1$ | 68.8 ± 0.6 | 75.9 ± 0.7 | 89.8 ± 1.3 | 68.4 ± 0.7 |
| $n = n_d$ | 70.6 ± 0.6 | 77.5 ± 0.3 | 92.2 ± 0.1 | 70.8 ± 0.2 |

Table 5: Target adaptation performance using softmax confidence based thresholding ($\tau$) and percentile based bagging ($\gamma$) schemes in our approach. Refer to Sec. 2.2b for discussion.

| Threshold | Office-31 ($\rightarrow$A) | ImageCLEF ($\rightarrow$P) | Office-Caltech ($\rightarrow$C) | Office-Home ($\rightarrow$Ar) |
|---|---|---|---|---|
| $\tau = 0.80$ | 70.4 ± 0.8 | 76.6 ± 0.4 | 92.4 ± 1.1 | 71.5 ± 0.5 |
| $\tau = 0.75$ | 70.8 ± 0.2 | 76.8 ± 1.0 | 93.3 ± 0.1 | 72.2 ± 0.3 |
| $\tau = 0.70$ | 71.8 ± 1.3 | 77.7 ± 0.4 | 92.8 ± 0.7 | 71.9 ± 0.4 |
| $\tau = 0.65$ | 72.6 ± 0.5 | 77.1 ± 0.7 | 92.4 ± 1.3 | 72.2 ± 0.5 |
| $\tau = 0.60$ | 71.4 ± 1.0 | 77.7 ± 0.8 | 92.5 ± 1.3 | 72.4 ± 0.5 |
| $\tau = 0.55$ | 73.4 ± 0.7 | 78.1 ± 0.5 | 93.1 ± 1.0 | 71.9 ± 0.2 |
| $\tau = 0.50$ | 72.5 ± 1.0 | 77.4 ± 0.9 | 92.3 ± 0.4 | 71.2 ± 0.1 |
| $\gamma = 5\%$ | 69.3 ± 0.2 | 77.4 ± 0.9 | 92.7 ± 0.3 | 73.2 ± 0.4 |
| $\gamma = 10\%$ | 70.3 ± 0.9 | 77.9 ± 0.2 | 92.8 ± 0.7 | 72.6 ± 0.3 |
| $\gamma = 15\%$ | 71.9 ± 0.8 | 78.6 ± 0.7 | 92.1 ± 0.3 | 72.8 ± 0.6 |
| SImpAl$_{50}$ | 70.6 ± 0.6 | 77.5 ± 0.3 | 92.2 ± 0.1 | 70.8 ± 0.2 |

samples for tuning appropriately. In our framework, we propose $w(\mathbf{x}_t, f, h)$, defined in Eq. 6 of the paper, as a measure of confidence in pseudo-label prediction for target instances, that results in an easy-to-hard curriculum (Sec. 4.2f in the paper). While our framework supports confidence based thresholding, we do not use it for the main results in the paper since it introduces additional hyperparameters. Here, we present empirical results by incorporating thresholds in SImpAl$_{50}$.

We incorporate two types of thresholds, softmax-confidence based (similar to Saito *et al.* [10]) and percentile-based bagging (top confident samples based on $w$). In both cases, target samples are first filtered based on classifier agreement, *i.e.* $\mathcal{D}_t' = \{(\mathbf{x}_t, y_t^{c_j}) \mid \mathbf{x}_t \in \mathcal{D}_t, \ a(\mathbf{x}_t, f, h) = 1\}$, and the samples are further chosen based on thresholding or bagging schemes.

For the softmax-confidence based threshold ($\tau$), we follow the method applied in Saito *et al.* [10], where a prediction is considered confident if at least one of the classifiers exhibit the $\arg\max$-confidence above a threshold $\tau$. Intuitively, self-training with such confident target samples will encourage more target samples to fall in the high confidence region. For the percentile-based bagging scheme, we choose the top $\gamma$-percentile (most confident) target samples from $\mathcal{D}_t'$ for pseudo-labeling. At each pseudo-label update, the percentile is increased by $\gamma$, *i.e.* the pseudo-labeled bag progressively grows in integral steps of $\gamma$. Both methods are trained for the same number of iterations as SImpAl$_{50}$.

Table 5 shows that self-training is sensitive to the confidence threshold $\tau$ and bagging percentile $\gamma$. Although the adaptation performance is seen to improve using thresholds, the best hyperparameter values are highly dataset specific. This calls for labeled target samples to reliably establish the most appropriate hyperparameter values for the task at hand. The study of automated methods to select threshold hyperparameters using unlabeled target samples would be of interest.

## 3 Practical Application: Cross-Domain Image Retrieval

We now demonstrate a practical application of Implicit Alignment. Consider an unlabeled dataset $\mathcal{D}$ of images sampled from a target domain. Suppose we wish to retrieve images from this dataset based on some class semantics (for *e.g.*, "retrieve images of objects having wheels"). With an unannotated dataset, this task seems non-trivial. However, we show that this is possible by performing Multi-Source Domain Adaptation on the target dataset $\mathcal{D}$ using SImpAl. Considering $\mathcal{D}$ as the target domain, we adapt a deep model using labeled source datasets and the unlabeled target data $\mathcal{D}$, under the SImpAl framework, and use this model to measure semantic similarity.

## Algorithm 1 Cross-Domain Image Retrieval

1: **require:** Reference image $\mathbf{x}_r$, Query set $\mathcal{D}$, Model $\{f, h\}$  ▷ $\cup$ denotes union

   ▷ Calculate the class probability vector for $\mathbf{x}_r$ (using Eq. 1 of the paper)
2: $\mathbf{M} \leftarrow h \circ f(\mathbf{x}_r)$
3: $\mathbf{p}_r \leftarrow \frac{1}{n_d} \sum_{i=1}^{n_d} \sigma(\mathbf{M}_{[i \cdot]})$

   ▷ Calculate class probability vectors for images in the query set $\mathcal{D}$ (using Eq. 1 of the paper)
4: $\mathcal{R} \leftarrow \{\ \}$ ▷ Create an empty collection of distance values
5: **for** $\mathbf{x}_q \in \mathcal{D}$ **do**
6: $\quad \mathbf{M} \leftarrow h \circ f(\mathbf{x}_q)$
7: $\quad \mathbf{p}_q \leftarrow \frac{1}{n_d} \sum_{i=1}^{n_d} \sigma(\mathbf{M}_{[i \cdot]})$
8: $\quad \mathcal{R} \leftarrow \mathcal{R} \cup \{||\mathbf{p}_q - \mathbf{p}_r||_2\}$ ▷ get the euclidean distance in class probabilities
9: $\mathcal{D}' \leftarrow$ set of $k$ nearest images in the query set $\mathcal{D}$ based on the distance $\mathcal{R}$

10: **return** $\mathcal{D}'$

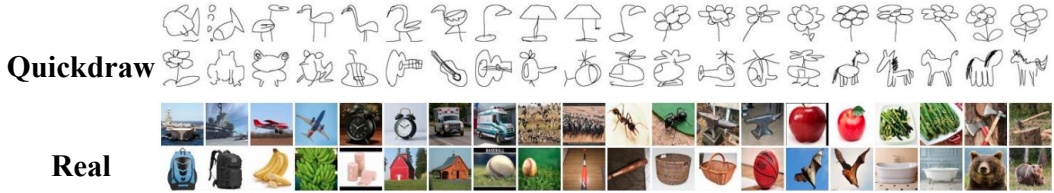

Figure 4: Sample images in the Quickdraw (**Qdr**) and Real (**Rel**) domains of the *DomainNet* [9] dataset. Note, this figure is meant for qualitatively demonstrating the domain-shift; the images do not correspond.

We develop a cross-domain image retrieval system where, given a reference image $\mathbf{x}_r$, we retrieve semantically similar images from a given set of images $\mathcal{D}$ (called the query set). To achieve this, we obtain the nearest neighbors of the reference image $\mathbf{x}_r$ in the set $\mathcal{D}$, based on the class probability vector $\mathbf{p}$ (as obtained in Eq. 1 of the paper). This is shown in Algorithm 1.

Note, Algo. 1 does not require any label information. Thus, the query set $\mathcal{D}$ can be unlabeled. We show an example use-case where we consider images in the "Quickdraw" (**Qdr**) domain of *DomainNet* [9] as the reference images, while the query set is the unlabeled target domain "Real" (**Rel**). Fig. 4 shows sample images in the two domains, which exhibit a large domain-shift.

We obtain a model adapted to the "Real" target domain (*i.e.* the model corresponding to →**Rel** in Table 1). Given a reference image in the **Qdr** domain, an end-user can retrieve semantically similar images from the unlabeled **Rel** domain using Algo. 1 above.

In Fig. 5, we present qualitative results, demonstrating the retrieval of images from the **Rel** domain using randomly selected images from the **Qdr** domain as reference. Note that, all images in Fig. 5 pertain to the test set of the corresponding domains that the model has not encountered during adaptation. The retrieval process using Algo. 1 above works surprisingly well, yielding qualitatively satisfactory results (annotated with green tick-mark). While there are certain images which yield false retrievals (annotated with red cross-mark), many of those cases have incomprehensible reference image (marked with an orange question-mark) and can be ignored during qualitative evaluation.

Further, we design a tool to retrieve images from a chosen query set, by manually "doodling" class semantics (similar to the images in the **Qdr** domain). See code implementation for the demonstration. Fig. 6 shows the images retrieved during the demonstration.

## 4 Code Reference

Pytorch implementation (with cross-domain image-retrieval demo) can be found on the project page[1].

---
[1] http://val.cds.iisc.ac.in/simpal

6

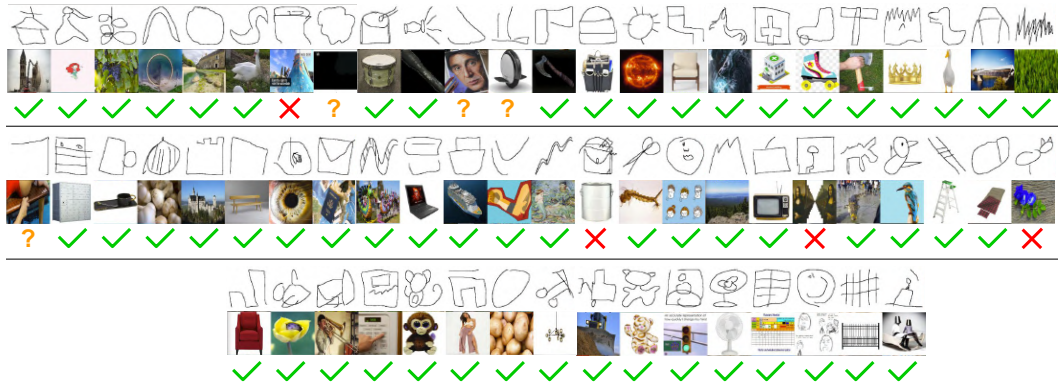

Figure 5: Images retrieved from the Real (**Rel**) domain (below) using reference images from the Quickdraw (**Qdr**) domain (above). The model corresponds to the →**Rel** task of *DomainNet*. Here, we show the nearest neighbor ($k = 1$ in Line 9 of Algo. 1 above). All images pertain to the test set of *DomainNet*. Green tick-marks indicate satisfactory retrivals, red cross-marks indicate false retrievals and orange question-marks indicate the cases where the reference image is incomprehensible. Note the high success rate of retrievals.

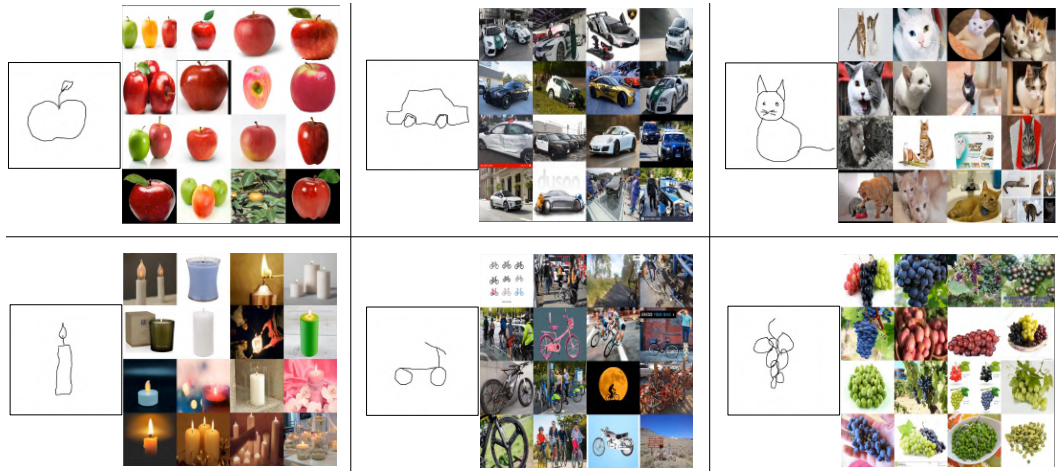

Figure 6: An example showing the top-16 retrievals ($k = 16$ in Line 9 of Algo. 1 above) from the **Rel** domain by manually "doodling" objects (see demonstration video on the project page).




\small

\printbibliography

\end{document}